\documentclass[letterpaper]{article}
\usepackage{aaai2027}
\usepackage[hyphens]{url}
\usepackage{graphicx}
\usepackage{natbib}
\usepackage{caption}
\usepackage{booktabs}
\usepackage{amsmath,amssymb}
\usepackage{algorithm}
\usepackage{algorithmic}
\title{Attacking Graph Foundation Models Through Their Shared Representation}
\author{Pankaj Kumar\textsuperscript{\rm 1 2}, Subhankar Mishra\textsuperscript{\rm 1 2}}
\affiliations{
\textsuperscript{\rm 1}{National Institute of Science Education and Research} \\
\textsuperscript{\rm 2}{An OCC of Homi Bhabha National Institute}}

\begin{document}
\maketitle

\begin{abstract}
    A graph foundation model generalizes across graph domains by mapping every input into one shared representation before any task reasoning. We call this map the alignment layer, the component that separates a graph foundation model from a graph neural network, and we show it is a distinct attack surface that prior work has not studied. We attack it at inference time, with no access to training, on six public models spanning spectral tokenizers, text embedding spaces, and a discrete codebook. A directed representation-space perturbation collapses every model, but at a budget comparable to the representation norm a plain graph network also needs, with one exception: OpenGraph, whose spectral tokenizer collapses at a fifth of that budget, an alignment-specific fragility a plain network does not share and which a same-representation control traces to the tokenizer rather than the decoder. A realizable input-space attack that edits edges, features, or text removes at least half the correct predictions on three of the six models at peak. How much of this fragility an input-access attacker realizes tracks how directly the decoder reads the representation, and not the clean accuracy a task leaves; we measure this carrier gain structurally from the decoder's local Lipschitz sensitivity, and report clean-accuracy headroom as a within-model ordering heuristic that does not survive on realizable attacks. Where the carrier is discrete we localize the effect causally by pinning the codebook assignment to its clean value. We then study defenses. Robustifying the representation fails against an adaptive attacker. Attacked inputs are, however, detectable: a density test on the low-dimensional continuous carriers separates them near perfectly, at a five percent false-positive rate on clean hold-out data, and holds under an adaptive attacker, so the alignment layer can be monitored even where our defenses do not harden it.
\end{abstract}

\section{Introduction}
\begin{figure*}[t]
\centering
\includegraphics[width=0.80\textwidth]{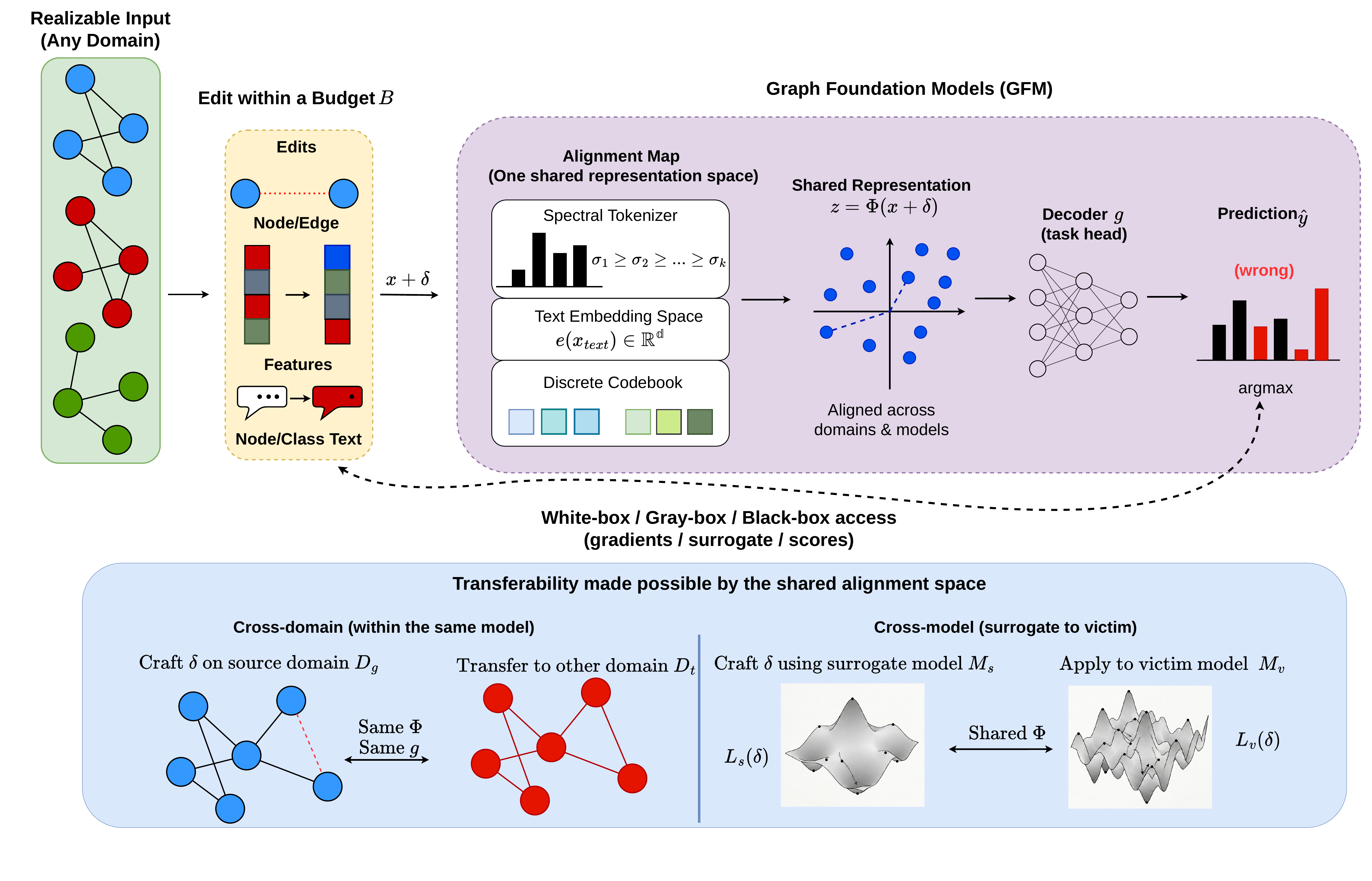}
\caption{Attack setup. A graph foundation model maps inputs from any domain into one shared
representation through the alignment map $\Phi$, read by a decoder $g$. We attack $\Phi$ at inference
time, editing a realizable input (edges, features, or node and class text) within a budget, with no
training access, under white-, gray-, or black-box knowledge. Because every domain and model family
shares this space, one perturbation transfers across domains and, through a surrogate, across models.
The carrier, the form $\Phi$ takes, is a spectral token, text embedding, codebook, or channel
logits.}
\label{fig:overview}
\end{figure*}

Graph foundation models aim to serve many graph tasks and many graph domains with one pretrained
model \citep{liu2023gfmsurvey,mao2024position}. Such a model must reconcile inputs that share no
nodes, no edges, and no feature space. It does so with a component that a graph neural network
does not have: an alignment layer that maps every input into one shared representation before
task reasoning begins. In structure models this layer is a spectral tokenizer built from the
singular value decomposition (SVD) of the adjacency
\citep{xia2024opengraph,xia2024anygraph,zhao2024graphany}. In text-attributed models it is a
frozen text embedding space or a discrete vocabulary
\citep{liu2024oneforall,wang2024gft,li2024zerog}. Figure~\ref{fig:overview} shows the common
structure: every domain is mapped into one shared representation, the attack perturbs that
representation, and the shared space carries the perturbation across domains and across models.

Adversarial attacks on graph neural networks perturb the adjacency or features of a single graph
for a single task \citep{zugner2018nettack,zugner2019metattack,xu2019topology,sun2019nipa}. None
can perturb a transferable encoder, an alignment map, or a discrete vocabulary, because those exist
only in a graph foundation model, to buy cross-domain generality. No published attack targets the
six we study.

One shared space means one shared weakness: because every domain routes through $\Phi$, a single displacement of it degrades them all, so the generality that makes these models foundational is itself the vulnerability. We carry the study from that vulnerability to a defense. Our contributions are as follows.
\begin{itemize}
\item We formalize the alignment map $\Phi$ shared across model families and attack it at
inference time, with no training access, on six public models (Section 3).
\item We separate the layer from the input lever with a plain-network control. Most alignment layers
collapse under a directed representation-space perturbation at the same relative budget as an
ordinary graph network, but OpenGraph's spectral tokenizer collapses at a fifth of that budget, a
fragility specific to the alignment layer, which a same-representation control traces to the
tokenizer rather than the decoder (Section 4).
\item We derive a spectral rotation attack for the singular-value tokenizer from eigenvector
perturbation theory, the first realizable attack to move OpenGraph off the noise floor where
projected gradient ascent fails, and show its residual resistance is spectral basis degeneracy, not
robustness (Section 4).
\item We evaluate the realizable input attack across nine datasets and four domains with multiple
seeds, and find that how much of the fragility an attacker realizes tracks how directly the decoder
reads the representation. We measure this carrier gain structurally, from the decoder's local
Lipschitz sensitivity at clean representations with no attack curve, and it rank-orders the collapse
threshold as a heuristic, while clean accuracy headroom does not survive as a
predictor on the realizable attacks (Sections 3 and 4). The discrete codebook admits a do-operator
that localizes the effect causally.
\item We show that robustifying the alignment layer fails against an adaptive attacker, but that
attacked inputs are detectable on the low-dimensional continuous carrier where robustification fails
(Section 5).
\end{itemize}
\section{Related Work}
\paragraph{Attacks on graph neural networks.} Structure attacks flip edges to change predictions.
Nettack crafts targeted edits \citep{zugner2018nettack}, Metattack poisons the graph with meta
gradients \citep{zugner2019metattack}, and topology attacks solve a min-max relaxation over edge
flips \citep{xu2019topology}. Reinforcement learning gives a black-box variant
\citep{dai2018adversarial}, and node injection adds nodes instead of editing edges
\citep{sun2019nipa}. Surveys cover the area \citep{jin2020adversarial}. All of this work assumes
one graph, one task, and one substrate.

\paragraph{Graph foundation models and their robustness.} Work on the robustness of graph
foundation models is recent and does not reach the alignment layer at inference. Benchmarks
perturb the prompt, text, and structure channels of graph language models on other models
\citep{zhang2025trustglm}, and backdoor attacks require a poisoning or fine-tuning stage
\citep{luo2025gfmba}. No prior attack shows a perturbation that transfers across domains through a
shared interface.

\paragraph{Representation alignment and certified robustness.} Independent models are argued to
converge toward a shared representation \citep{huh2024platonic}, but a representation can be
decodable without being used causally \citep{convergence2026}, which is why our test intervenes
rather than probes. A representation-space perturbation transfers across models only when
representations are geometrically aligned \citep{gupta2025advtransfer}, the condition a cross-domain
alignment map creates, so this attack surface is also a transfer surface. We identify subspaces with
linear alignment measures \citep{kornblith2019cka} and remove them with erasure and activation
editing \citep{belrose2023leace,meng2022rome}. Perturbing an internal representation is an
established paradigm: feature adversaries match a target representation
\citep{sabour2016manipulation}, the idea reaches vision-language and self-supervised encoders
\citep{zhao2023vlmrobust,jia2022badencoder}, and matching a target in a shared embedding space
transfers across encoders \citep{zhang2024illusions}. There the contribution is the target, not the
optimizer, and an unrealizable representation-space perturbation is the accepted way to show a
vulnerability before asking whether an input can reach it. We move the target to the representation a
graph foundation model adds over a graph network: the cross-domain alignment map, not one model's
private features. Our defense adapts randomized smoothing \citep{cohen2019smoothing} with adaptive
evaluation following standard practice
\citep{carlini2017towards,athalye2018obfuscated,tramer2020adaptive}.

\section{The Alignment Map and the Attack}
Let a graph foundation model read an input $x=(A,X,\text{text})$ from any domain, where $A\in\{0,1\}^{n\times n}$ is the adjacency, $X\in\mathbb{R}^{n\times d}$ the node features, and $\text{text}$ the node and class descriptions. Define the
alignment map $\Phi:(A,X,\text{text})\to S$ as the representation every input passes through
before task reasoning, so the model is $g\circ\Phi$, and let $S$ denote this shared representation space (the image of $\Phi$). Its image is shared across domains by
construction. An alignment layer is not any encoder. It has four properties. It is
domain-agnostic: the same map serves every input domain, not one encoder per domain. It is a
bottleneck: every input passes through it before task reasoning. It is a low-dimensional or
discrete object: a top-$k$ singular subspace, a codebook, or a fixed embedding space. It is read
by a task decoder $g$. An intermediate layer of a single-domain graph neural network is not an
alignment layer, because it is not shared across domains and supports no cross-domain transfer.
Because $\Phi$ is one cross-domain bottleneck, a single perturbation of $S$ moves every domain
that routes through it, so it can express cross-domain transfer. An attack on an ordinary hidden
layer is bound to one input distribution and cannot express such a transfer. For a spectral model, with $D$ the diagonal degree matrix, $\bar A = D^{-1/2}AD^{-1/2}$ the symmetrically normalized adjacency, and $U\Sigma V^\top$ its singular
value decomposition ($U,V$ the singular vectors, $\Sigma$ the diagonal of singular values $\sigma_1\ge\sigma_2\ge\cdots$), the tokens are
\begin{equation}
E = \Big(\textstyle\sum_{l=1}^{L}\bar A^{l}\Big)\,\mathrm{LN}\big(U\sqrt\Sigma + V\sqrt\Sigma\big),
\label{eq:tok}
\end{equation}
where $\mathrm{LN}(\cdot)$ is row-wise layer normalization, and $S$ is the leading singular subspace of $\bar A$. For a text model $S$ is a frozen text
embedding space, and predictions read the similarity between a node embedding and a class
embedding. For GFT $S$ is a discrete codebook $C$, and a computation-tree embedding $z$ is
assigned the nearest token, $j=\arg\min_c \lVert z-c\rVert$. We make two claims about $\Phi$ and one
hypothesis, and Section 4 tests each.

\noindent\textbf{Claim 1 (shared surface).} $\Phi$ is a bottleneck that every domain passes
through. Attacking it is therefore a different operation from attacking message passing, which is
tied to one adjacency and one task, with no analogue in the GNN setting.

\noindent\textbf{Claim 2 (cross-domain carrier).} Because $\Phi$ maps every domain into one
geometry, a displacement inside $S$ can transfer to a target domain that shares $\Phi$, which a structure attack on a single, non-shared adjacency cannot express.

\noindent\textbf{Hypothesis (concentration).} $\Phi$ compresses the model into a low dimensional
object: a top-$k$ singular subspace, a codebook, or a routed expert, so a small displacement of $S$
changes the output through a basis rotation, a token flip, or an expert re-route. Whether this
concentration is what makes a layer fragile is what we test, and Section 4 finds that the collapse
threshold orders against it, so we report it as a hypothesis the evidence does not support.

\paragraph{Why the alignment map is the surface.}
Write the model as $g\circ\Phi$, where $\Phi$ produces a rank-$k$ representation in $S$ and $g$
is an $L$-Lipschitz decoder. Two facts make $\Phi$ fragile. First, a bounded input budget $B$
rotates the leading singular subspace of $\bar A$ by an amount set by the inverse singular gap
$1/(\sigma_i-\sigma_j)$ (Davis-Kahan), so a near-degenerate spectrum turns a small budget into a
large displacement inside $S$. Second, only the part of that displacement the decoder reads
changes the output, so a trained decoder with a small task-subspace absorbs the rest. These give a
flip-budget proposition with two factors, a margin and a carrier gain.

\smallskip
\noindent\textbf{Proposition (flip budget).}
\emph{Let node $i$ have clean margin $m_i>0$ (the signed distance of node $i$ to its decision boundary in $S$), and let the attack move its representation with carrier gain $\kappa_i>0$ (the boundary-normal displacement per unit budget). To first order the minimal budget to flip node $i$ is $b_i=m_i/\kappa_i$, so reachability is $R(B)=\Pr[\,m_i\le\kappa_i B\,]$.} We show in Section 4 that on realizable attacks the carrier gain
$\kappa$ dominates, and the margin term, which clean accuracy summarizes, does not order
reachability on its own. Proofs are in the supplement.

\begin{algorithm}[t]
\caption{ALIGN attack (task-loss form)}
\label{alg:align}
\textbf{Input}: input $x$, model $g\circ\Phi$, labels $y$, budget $B$, steps $T$\\
\textbf{Output}: perturbed input $x'$
\begin{algorithmic}[1]
\STATE initialize $\delta \gets 0$
\FOR{$t=1$ to $T$}
\STATE $s \gets \Phi(x+\delta)$ \quad (fixed-basis surrogate if the SVD is unstable)
\STATE $\ell \gets \mathrm{loss}(g(s), y)$
\STATE $\delta \gets \delta + \eta\,\nabla_\delta \ell$
\STATE project $\delta$ onto the budget $B$
\ENDFOR
\STATE \textbf{return} $x' = x + \delta$ projected to the discrete input space
\end{algorithmic}
\end{algorithm}

\paragraph{Threat model.} The attacker acts at inference time only, with no access to training,
poisoning, or fine-tuning, and no ability to modify model weights (Figure~\ref{fig:overview}).
Three axes vary. \emph{Goal:} untargeted evasion, with a targeted codebook variant in the ablations.
\emph{Capability:} what is perturbed per carrier (edges and SVD features, input features, node or class text, or the pre-quantization embedding), each within a stated budget. An attack is \emph{realizable} when its perturbation is an input the model actually accepts; a representation-space perturbation the attacker cannot emit, such as the codebook embedding, is reported only as an upper bound.
\emph{Knowledge:} white box, gray box (surrogate, no victim gradients), or black box (scores or transfer only). We report each capability separately and do not combine them. The realizable input attacks impose no unnoticeability constraint beyond the stated budget, so their numbers upper-bound attacker power. Budget definitions per carrier and two fidelity caveats (OFA's decoder, ZeroG's pipeline) are in the supplement.

\paragraph{The ALIGN attack.} The objective is a small perturbation $\delta$, within a budget,
such that $g(\Phi(x+\delta))$ is incorrect or takes a target label. We instantiate it per
carrier. For a spectral tokenizer we perturb edges, and features where they enter the SVD, to
rotate the leading singular subspace. That objective has zero gradient at the clean graph, so we
maximize the task loss and treat the rotation as a consequence, with a fixed-basis surrogate where
differentiating the SVD is unstable. For a text space we perturb node text or the shared
class-description text, since one class embedding is scored against every node. For the codebook we
push a computation-tree embedding across the boundary to a target token with a straight-through
estimator. The general optimizer is task-loss projected gradient ascent on $g\circ\Phi$, and what
the alignment view contributes is which component to perturb and a perturbation that transfers
across domains. In a budget-matched comparison a generic end-to-end perturbation is stronger on its
source domain, but one confined to the shared subspace transfers several times better across a real
domain gap (supplement).

\paragraph{Align-then-ablate.} To test whether the attack is carried by the alignment layer we
intervene rather than probe. We identify the alignment subspace $P$ by fitting an orthogonal
Procrustes map between paired cross-domain representations and taking the rank at which their
similarity peaks (supplement), remove it by projection, and report the mediated fraction
$1-(A^{\text{abl}}_{\text{clean}}-A^{\text{abl}}_{\text{atk}})/(A_{\text{clean}}-A_{\text{atk}})$
against a random subspace of the same rank, over a range of ranks. For the discrete codebook the
intervention is a do-operator that pins the token index to its clean value.

\section{Experiments}
We reproduce each model to its published clean number before attacking, then attack across each
model's zero-shot suite with multiple seeds. Table~\ref{tab:main} and Figure~\ref{fig:main}(a)
summarize the result. The full seeded matrix is in the supplement.

\begin{figure*}[t]
\centering
\includegraphics[width=0.75\textwidth]{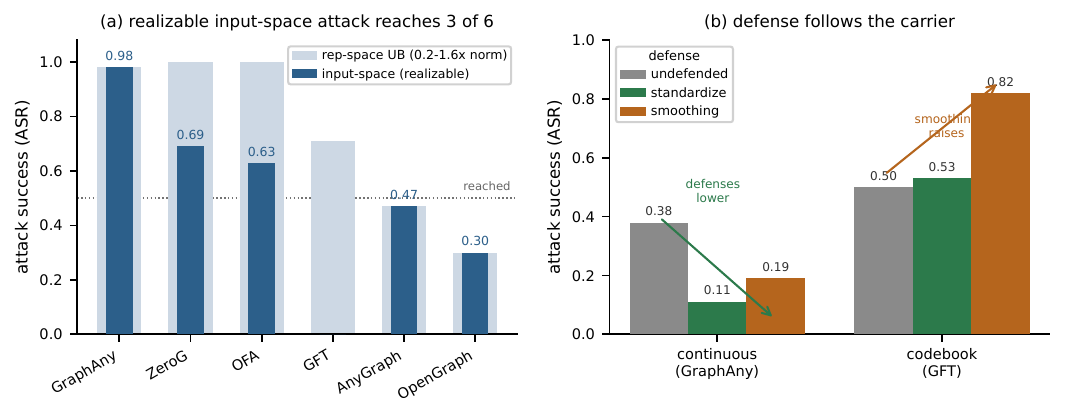}
\caption{Left: input-space attack success reaches three of six models above one half, with the
representation-space peak drawn as a lighter upper bound the attacker cannot emit, and reached only
at a directed budget of $0.2$ to $1.6$ of the representation norm, not commensurable with the input.
An equal budget random perturbation stays near zero except on the densest feature budgets. Right: the defense
follows the carrier. Standardization and smoothing lower the attack on the continuous carrier, while
smoothing at $\sigma{=}0.5$ raises it on the codebook to $0.82$ because the added noise flips tokens.
The defense table reports the codebook at $\sigma{=}0.25$.}
\label{fig:main}
\end{figure*}

\begin{table}[t]\centering\small
\setlength{\tabcolsep}{4pt}
\begin{tabular}{llccccc}
\toprule
Model & Carrier & $\kappa$ & In.\ peak & In.\ mean & Rep.\ UB \\
\midrule
GraphAny & features & 7.1 & \textbf{0.98} & 0.76 & 0.99 \\
ZeroG    & text     & 5.9 & \textbf{0.69} & 0.32 & 1.00 \\
OFA      & text     & 5.2 & \textbf{0.63} & 0.59 & 1.00 \\
GFT      & codebook & -- & -- & -- & 0.71 \\
AnyGraph & SVD edge & 2.5 & 0.47 & 0.31 & 0.82 \\
OpenGraph & SVD edge & 0.3 & 0.30 & 0.17 & 0.99 \\
\bottomrule
\end{tabular}
\caption{Main evasion result. Carrier gain $\kappa$ is a post-hoc descriptor, the initial slope of
the attack-success-versus-budget curve in each carrier's own unit, comparable only within a carrier;
GFT has no entry, lacking a realizable input attack. Columns then give the input-space attack success
(peak and mean over a model's datasets) realizable by the stated threat model, and the
representation-space upper bound from perturbing the alignment representation directly, which the
input lever cannot reach. The saturated bound is high for every model, but a plain network collapses
at a similar relative budget (Section 4) except on OpenGraph. The input-space column reaches three
models at peak and two by mean (bold). Full seeded matrix in the supplement.}
\label{tab:main}
\end{table}

\paragraph{Setup.} We evaluate two spectral tokenizer models, OpenGraph and AnyGraph, a closed-form
spectral-filter model, GraphAny, two text-space models, OFA and ZeroG, and one codebook model, GFT,
each reproduced to its published clean number and wrapped behind one interface so the attack code is
shared. The datasets span citation, web, e-commerce, heterophily, and social domains, together with
AnyGraph's native link-prediction graphs on which it is evaluated in its own zero-shot setting (full
list in the supplement). Attacks run on each model's real inference pipeline. Where a white-box
gradient passes through a non-differentiable step, such as the entropy-normalized distance in
GraphAny or the codebook argmax in GFT, we craft with a torch surrogate and re-evaluate every number
on the real pipeline. Budgets are a fraction of the edges or an $\ell_\infty$ or $\ell_0$ bound on
features, each cell averaging over seeds, against an equal-budget random control. Attack success is
the relative accuracy degradation $(A_{\rm clean}-A_{\rm atk})/A_{\rm clean}$, or the relative
Recall@20 drop for AnyGraph. We report the untargeted attack unless noted, and call a model reached
when its peak attack success exceeds one half.

\paragraph{Three models are reached by a realizable input-space attack.} GraphAny's feature attack,
which edits the input features, succeeds across all four of its domains (the full matrix (supplement)). A
sparse variant that changes about four feature entries per node already halves accuracy, and the
random control stays near zero, so the strength is the direction, not the budget. The realizable
text attack on OFA and ZeroG edits the shared class-description words under a semantic-similarity
constraint, and one edit moves the score of every node in that class. It reaches attack success
above one half on the citation graphs but zero on the two social graphs, where no improving edit
exists (Table~\ref{tab:main}). GFT's codebook attack succeeds and holds up to ogbn-arxiv, but it is
a representation-space perturbation with no realized input-space form, so we report it as an upper
bound.

\paragraph{OpenGraph's alignment layer is fragile beyond a plain network, the others are not.} To
separate the lever from the layer we perturb the representation the decoder reads directly, at a
directed budget in units of the representation norm. Plain single-domain GCN, GraphSAGE, and GAT
collapse at a fraction $0.48$ to $0.70$ of the norm (mean $0.61$), and GraphAny ($0.62$) and AnyGraph
($0.56$) sit in this band, so their representation-space collapse is generic. OpenGraph is the
exception: its spectral token collapses at $0.12$, five times below that band, and a
same-representation control (a linear probe and a perceptron on its own clean token collapse at
$0.04$ and $0.09$, at or below the transformer's $0.15$) locates the cause in the tokenizer, not the
decoder: the singular value decomposition produces a large-norm, small-margin token any classifier
inherits (supplement).

\paragraph{A spectral rotation attack reaches OpenGraph through edges.} Projected gradient ascent
sits at the noise floor on the spectral tokenizer, since the singular basis is detached from the
gradient. An edge attack derived from first-order eigenvector perturbation theory recovers the
subspace rotation and moves OpenGraph off the noise floor, to attack success $0.15$ at a tenth of the
edges, five times a random flip; the residual resistance is spectral basis degeneracy, not
robustness (supplement).

\begin{figure}[t]
\centering
\includegraphics[width=\columnwidth]{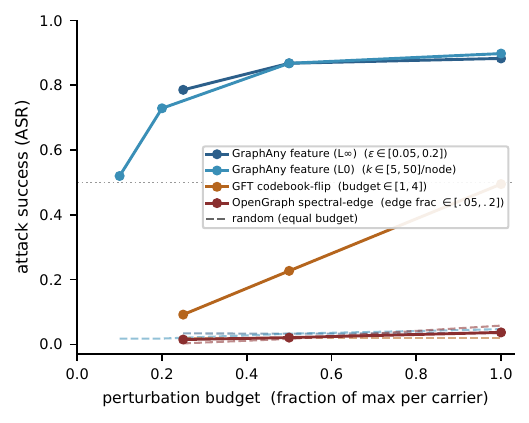}
\caption{Attack success against budget for a feature carrier (GraphAny), a codebook carrier (GFT),
and a spectral edge carrier (OpenGraph) on Cora, with matched equal-budget random controls dashed.
Budget is normalized to a fraction of its per-carrier maximum, since the units differ. The carrier
gain $\kappa$ is the initial slope of these curves.}
\label{fig:budget}
\end{figure}

\begin{figure}[t]
\centering
\includegraphics[width=\columnwidth]{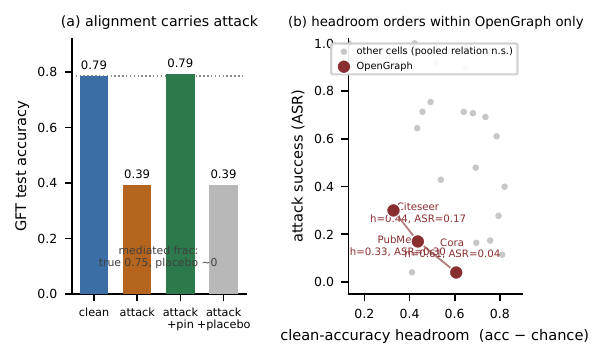}
\caption{Left: removing the alignment subspace collapses the attack on GFT, while a random
subspace of the same rank does not, so the alignment layer carries the attack. Right: within
OpenGraph the three datasets order by headroom, lower headroom giving higher attack success, but
this within-model pattern does not survive pooling across models, where the relation is not
significant.}
\label{fig:causal}
\end{figure}

\paragraph{What governs the input lever.} Headroom is the clean accuracy a task leaves above chance,
$h=A_{\rm clean}-1/C$ for $C$ classes. Within a single model lower headroom can order reachability,
as OpenGraph does across Cora, Citeseer, and PubMed (Figure~\ref{fig:causal}(b)), but pooled across
the twenty-one realizable cells it does not (Spearman $0.01$, $N=21$, supplement). A decoder-side
structural predictor does better: the local Lipschitz constant $L$ of the decoder gives a collapse
threshold $m/(L\,\|S\|)$ per node with no attack curve, dimensionless where the fitted $\kappa$ and
the input-side Davis-Kahan gap are not. It rank-orders the measured threshold across models and
datasets (Spearman $0.65$, $N=11$) and reproduces OpenGraph's threshold from clean representations
alone, though not with OpenGraph held out ($0.53$), so we read it as a heuristic, not a law
(supplement). We report the fitted $\kappa$ only within a carrier (Spearman $0.63$ to $0.87$).
Figure~\ref{fig:budget} plots attack success against budget for one carrier of each type. The attack
also reaches auxiliary surfaces (node injection, router re-routing), and combining input channels
does not beat the strongest single lever (supplement).

\paragraph{Cross-domain transfer is selective.} A single perturbation along a generic axis of the
shared text space, crafted on one citation graph, transfers to some targets and not others
(supplement). The sharper evidence is the budget-matched comparison in Table~\ref{tab:e2e}: one
universal vector crafted on Cora, applied unchanged to near-domain Citeseer and far-domain PubMed,
either free in the full embedding or confined to the top sixteen shared directions. The generic
vector is stronger on the source, because about half its budget lies outside the shared subspace and
does not travel. On far-domain PubMed it reaches attack success $0.06$, matching a random control,
while only the shared-subspace vector transfers above chance, to $0.22$ (three seeds, standard
deviation below $0.005$). The gap is small but stable, and it is the one operation attacking an
encoder cannot express: a universal adversarial perturbation \citep{moosavi2017universal} carried
through the alignment layer, where confining it to the shared low-dimensional subspace is what lets
one vector fool a second domain. Citeseer does not distinguish the two conditions, its clean
accuracy ($0.27$) being at chance.

A gray-box variant, where an edge set crafted
on one spectral model is applied to another with no gradient access to it, also transfers above a
random control at larger budgets though the absolute effect stays small on the resistant spectral
pair (supplement). A black-box variant that reads only output scores, with no gradients, recovers
most of the white-box effect on the text carrier at a few hundred queries per node (supplement),
so the threat model is evidenced at all three levels of access.

Three factors govern transfer: (i) Perturbation must lie along a generic content axis; an
attack-tuned universal vector overfits the source and does not travel, (ii) Decoder's
normalization decides exposure: ZeroG standardizes its node embeddings before the readout, which
cancels a shared additive shift and defends it, while OFA does not and is exposed, and (iii) Transfer weakens with domain distance. This points to per-domain normalization as a cheap transfer defense.

\paragraph{Causal localization.} Pinning GFT's codebook index to its clean value under attack
restores accuracy, a do-operator on the discrete carrier that holds across GFT's node suite
(Figure~\ref{fig:causal}(a)). For a continuous spectral carrier the effect does not concentrate
in the leading subspace, so the discrete codebook gives the clean causal evidence.

\subsection{Results by Model}
The carrier type organizes the outcome (Table~\ref{tab:main}, per-dataset in
the full matrix (supplement)). Where a light decoder reads the alignment representation the attack
succeeds. Where a trained network sits between the representation and the output the same move is
absorbed unless the task is fragile.
\paragraph{GraphAny.} Its channels are closed-form solutions read by a light fusion, so a feature
perturbation that moves the solved logits changes the output. It succeeds across all four domains at
both a dense and a sparse budget of a few entries per node (supplement), and the random control does
not move accuracy. What is specific to the foundation model is that the solve runs over a fixed
spectral basis shared across every domain with no per-graph training, so one perturbation moves the
zero-shot path the model reuses on every dataset.

\paragraph{OpenGraph.} A trained transformer reads the tokenizer and absorbs a generic edge
perturbation where clean accuracy is ample. It succeeds once the task is fragile, so within
OpenGraph headroom orders reachability, a within-model pattern that does not survive pooling
(Figure~\ref{fig:causal}(b)).
\begin{table}[t]\centering\small
\setlength{\tabcolsep}{4pt}
\begin{tabular}{llrrr}
\toprule
Budget ($\ell_2$) & Search space & Cora & Citeseer & PubMed \\
\midrule
1.34 & generic (full) & 0.56 & 0.34 & 0.02 \\
1.34 & shared subspace & 0.34 & 0.34 & 0.08 \\
2.68 & generic (full) & 0.76 & 0.34 & 0.06 \\
2.68 & shared subspace & 0.51 & 0.34 & \textbf{0.22} \\
\bottomrule
\end{tabular}
\caption{One universal perturbation crafted on Cora at matched budget, attack success on each
target, mean over three seeds with standard deviation below $0.005$. The generic vector wins on the
source. The shared-subspace vector transfers to far-domain PubMed, the differentiator of the
alignment framing.}
\label{tab:e2e}
\end{table}

\paragraph{AnyGraph.} The edge attack on its joint tokenizer succeeds on several datasets, and being
in-distribution for the checkpoint gives no protection (supplement). Its feature half is inert, so
the edge channel carries the attack.

\paragraph{OFA and ZeroG.} Both read a similarity between a node embedding and a class embedding in
one text space. Editing the shared class-description text is the realizable attack, and one edit
moves every node in that class, reaching attack success above one half on the citation graphs but
zero on the two social graphs, where a two-class decoder admits no improving edit
(Table~\ref{tab:main}). Perturbing the node embedding directly is higher but an upper bound the
attacker cannot emit.

\paragraph{GFT.} The codebook is discrete, so the attack flips token assignments, and pinning the
tokens to their clean value restores accuracy. This gives the clean causal evidence
(Figure~\ref{fig:causal}(a)), and the effect holds up to ogbn-arxiv, about 170k nodes.

\paragraph{Classical attacks cover only the structural surface.} A classical graph attack perturbs
the adjacency, so it is defined only where a model reads one, limiting the comparison to OpenGraph
node classification. There Metattack and PGD transferred from a surrogate graph convolutional
network exceed our tokenizer-targeted attack on Cora and Citeseer and beat it on fragile PubMed
(supplement), so we claim no advantage on the structural surface. The point is the reverse. Nettack
edits one graph for one task \citep{zugner2018nettack}, with no form that shifts a class embedding
read against every node, flips a shared codebook token, or carries a perturbation across domains
through the alignment map. The text carrier, the codebook, and the cross-domain transfer have no
classical counterpart.

\paragraph{Ablations.} The attack objective matters (A1): the task loss and its fixed-basis
surrogate reach high success while the subspace rotation objective does not, because its gradient
vanishes at the clean graph. The causal effect is graded in the ablated rank on GFT (A3), while
continuous carriers do not localize. Spectral degeneracy lowers the attack cost within a graph
(A5), and a targeted codebook flip is more damaging than an untargeted one although it flips
fewer tokens (A9). Full tables are in the supplement.

\section{Defense: Not Correctable, but Detectable}
\paragraph{Robustifying the alignment layer fails.} We derive four defenses from the causes above,
standardization, smoothing, margin hardening, and an off-the-shelf spectral purification, and
evaluate each against an adaptive attacker, which is the correct test. None survives (appendix).
Standardization is the strongest static defense but a defense-aware attacker differentiates through
it and recovers the full attack. Smoothing gives a certified radius in carrier space that does not
cover the attack budget \citep{cohen2019smoothing,bojchevski2020sparse}, and an attacker that
averages over the noise recovers most of its benefit. On the codebook the noise these defenses add
itself flips tokens, so they backfire even statically. GCN-SVD is unusable on the spectral carrier,
because OpenGraph's tokenizer is itself a truncated singular value decomposition of the same
adjacency \citep{entezari2020svd}, so the subspace it keeps is the one the attack rotates. A decoder
that resists a single token flip removes that one lever, but an adaptive attacker moves to the
continuous prototype head and total attack success rises above the undefended level
(Table~\ref{tab:defmain}).

\begin{table}[t]\centering\small
\setlength{\tabcolsep}{4pt}
\begin{tabular}{lcccc}
\toprule
Defense (carrier) & Clean$_{\text{def}}$ & Undef. & Static & Adapt. \\
\midrule
standardize (GraphAny) & 0.78 & 0.38 & \textbf{0.11} & 0.39 \\
smoothing (GraphAny)   & 0.80 & 0.38 & 0.19 & 0.31 \\
smoothing (GFT codebook) & 0.78 & 0.49 & 0.68 & 0.72 \\
GCN-SVD (OpenGraph) & 0.42 & 0.15 & 0.06$^{\dagger}$ & -- \\
\bottomrule
\end{tabular}
\caption{Attack success under each defense, static and adaptive. None survives the adaptive
attacker, on the codebook they backfire even statically, and GCN-SVD is unusable (the
$^{\dagger}$ drop is clean accuracy collapsing, undefended clean $0.61$). Full table in the
supplement.}
\label{tab:defmain}
\end{table}

\paragraph{Attacked inputs are detectable.} Flagging the attack does not require correcting it. The
attack pushes the representation off the clean manifold, so a density test fit on clean
representations alone separates attacked from clean inputs, almost perfectly on the low-dimensional
continuous carriers: a Mahalanobis test on GraphAny's fusion distances and GFT's pre-quantization
embedding reaches AUC $1.00$ at a five percent false-positive rate. It holds under an adaptive
attacker adding a stay-on-manifold penalty (GFT stays near AUC $0.87$; on GraphAny the attacker pulls
it toward chance only by cutting attack success from $0.37$ to $0.12$), and flags a
feature-distribution attack \citep{inkawhich2020transferable} where that attack flips the prediction
(supplement). The test is fit on clean representations only, and is not uniform, falling to $0.66$ on
ZeroG's near-isotropic text embedding, so it tracks the carrier's effective dimension
(Table~\ref{tab:detmain}), the graph analogue of an image result that a perturbation raises the local
intrinsic dimensionality \citep{ma2018lid}.

\begin{table}[t]\centering\small
\setlength{\tabcolsep}{4pt}
\begin{tabular}{lccccc}
\toprule
Model & Carrier & Eff.\ dim & AUC & TPR@5\% & Adapt. \\
\midrule
GraphAny & fusion & 5 & 1.00 & 1.00 & 0.74 \\
GFT & codebook $z$ & 3 & 1.00 & 1.00 & 0.88 \\
ZeroG & text & 73 & 0.66 & 0.08 & 0.92 \\
\bottomrule
\end{tabular}
\caption{Detection of attacked inputs by a density test fit on clean representations. Effective
dimension is the participation ratio of the clean representation, which tracks detectability, not
the ambient dimension. GFT's $768$-dimensional embedding sits near a discrete codebook, so its
effective dimension is $3$ and it is caught, while ZeroG's near-isotropic text embedding has
effective dimension $73$ and is not.}
\label{tab:detmain}
\end{table}

\section{Discussion and Limitations}
On the structure-only surface of OpenGraph, classical attacks transfer better than ours; the
contribution there is the alignment surfaces they cannot express and the causal localization they do
not provide. Causal localization is clean for the discrete codebook but confounded for continuous
carriers, where no linear erasure (LEACE, INLP) both preserves clean accuracy and removes the
effect, so the discrete do-operator is the only clean handle. The certified radius is in carrier
space; an input-space guarantee needs a Lipschitz bound on $\Phi$, which the codebook lacks, since
nearest-token assignment is piecewise constant, so the certificate cannot lift to the input there.
The capabilities differ in realism: text edits are realistic when the attacker supplies content and
we constrain them by semantic similarity, while feature and edge perturbations assume control of
attributes or links and impose none of the unnoticeability constraints standard in the graph-attack
literature \citep{zugner2018nettack}, so their numbers upper-bound attacker power.

\section{Conclusion}
The alignment layer that maps every domain into one shared space is an attack surface that has not
been studied. Most alignment layers are no more fragile under a directed representation-space
perturbation than a plain graph network, but OpenGraph's spectral tokenizer collapses at a fifth of
the budget, a fragility specific to the alignment layer that a carrier-specific edge attack
partially realizes. An input-access attacker breaks three of the six, gated by how directly the
decoder reads the representation, not by clean accuracy. Robustifying the layer fails against an
adaptive attacker, but attacked inputs are detectable on its low-dimensional carriers.

\section{Ethical Statement}
This is robustness research on public benchmarks and released models. All attacks are
inference-time evasion, reported so that builders can defend the alignment layer. We target no
deployed system and release code for reproducibility.

\bibliography{main}

\appendix
This supplement contains the full seeded results matrix (\S\ref{s:matrix}), the classical-attack
baseline comparison (\S\ref{s:base}), cross-domain transfer numbers (\S\ref{s:transfer}), the
complete ablation battery (\S\ref{s:abl}), the robust-GFM tables including the certified-robustness
and joint-retrain results (\S\ref{s:rgfm}), and method, dataset, and reproduction details
(\S\ref{s:method}).

\section{Full Evasion Matrix}
\label{s:matrix}
Table~\ref{tab:fullmatrix} reports the peak attack success rate (relative accuracy drop) for each
model and dataset, best over our attacks and the reported budget sweep. Random-perturbation
baselines of equal budget are near zero throughout (0.00--0.03), except on the densest-feature
graphs (PubMed, WikiCS) where even random noise moves accuracy somewhat. Cells derived from a
seeded sweep report the mean. The per-seed rows are in the released \texttt{results.json}
(\texttt{evasion\_seeds}). AnyGraph rows are flagged for the \texttt{link2}-pretrained checkpoint:
Cora and CS are in-distribution (seen), only Citeseer, PubMed, products, and p2p are genuine
zero-shot.

\begin{table*}[t]\centering\small
\setlength{\tabcolsep}{6pt}
\begin{tabular}{llrrll}
\toprule
Model & Dataset & Peak ASR & $n$ & Best attack & Space \\
\midrule
AnyGraph & CS & 0.11$\pm$0.04 & 3 & spectral-edge & input \\
 & Citeseer & 0.26$\pm$0.05 & 3 & spectral-edge & input \\
 & Cora & 0.40$\pm$0.02 & 3 & spectral-edge & input \\
 & PubMed & 0.17$\pm$0.01 & 3 & spectral-edge & input \\
 & p2p-Gnutella06 & 0.43$\pm$0.08 & 3 & spectral-edge & input \\
 & products\_home & 0.47$\pm$0.00 & 3 & spectral-edge & input \\
GFT & Cora & 0.49$\pm$0.03$^{\ast}$ & 3 & codebook & rep \\
 & PubMed & 0.04$\pm$0.03$^{\ast}$ & 3 & codebook & rep \\
 & WikiCS & 0.16$\pm$0.04$^{\ast}$ & 3 & codebook & rep \\
 & ogbn-arxiv & 0.71$\pm$0.03$^{\ast}$ & 3 & codebook & rep \\
GraphAny & Amazon-Computers & 0.69$\pm$0.02 & 3 & feature-L0 & input \\
 & Amazon-Photo & 0.61$\pm$0.02 & 3 & feature-L0 & input \\
 & Citeseer & 0.92$\pm$0.00 & 3 & feature-Linf & input \\
 & Cora & 0.87$\pm$0.00 & 4 & feature-L0 & input \\
 & Cornell & 0.71$\pm$0.05 & 3 & feature-Linf & input \\
 & PubMed & 0.98$\pm$0.00 & 3 & feature-Linf & input \\
 & Texas & 0.43$\pm$0.05 & 3 & feature-L0 & input \\
 & WikiCS & 0.98$\pm$0.00 & 3 & feature-L0 & input \\
 & Wisconsin & 0.64$\pm$0.09 & 3 & feature-L0 & input \\
OFA & Cora & 0.96$\pm$0.00$^{\ast}$ & 4 & node-emb & rep \\
 & Pubmed & 1.00$\pm$0.00$^{\ast}$ & 3 & node-emb & rep \\
OpenGraph & Citeseer & 0.17$\pm$0.02 & 3 & spectral-edge & input \\
 & Cora & 0.04$\pm$0.01 & 3 & spectral-edge & input \\
 & PubMed & 0.30$\pm$0.02 & 3 & spectral-edge & input \\
ZeroG & Citeseer & 1.00$\pm$0.00$^{\ast}$ & 3 & node-emb & rep \\
 & Cora & 0.75$\pm$0.02$^{\ast}$ & 4 & class-emb & rep \\
 & Pubmed & 1.00$\pm$0.00$^{\ast}$ & 3 & node-emb & rep \\
 & instagram & 1.00$\pm$0.00$^{\ast}$ & 3 & node-emb & rep \\
 & reddit & 1.00$\pm$0.00$^{\ast}$ & 3 & node-emb & rep \\
\bottomrule
\end{tabular}

\caption{Full evasion matrix. Peak ASR per model and dataset.}
\label{tab:fullmatrix}
\end{table*}

\paragraph{GraphAny, per dataset (three seeds).} Citation: Cora $0.87\pm.00$, Citeseer
$0.92\pm.00$, PubMed $0.98\pm.00$. Web: WikiCS $0.91\pm.01$. E-commerce: Amazon-Photo
$0.61\pm.04$, Amazon-Computers $0.68\pm.03$. Heterophily: Cornell $0.71\pm.05$, Texas
$0.39\pm.05$, Wisconsin $0.55\pm.05$. The $\ell_0$ sparse-feature variant reaches ASR 0.52 at
$k{=}5$ ($\sim$4 entries/node) and matches the dense $\ell_\infty$ attack (0.87) at $k{=}25$,
while a random-$\ell_0$ control stays at 0.02--0.05.

\paragraph{Scale.} The attack is not limited to small citation graphs. It is evaluated on graphs
from 2.7k to 169k nodes across citation, wiki, e-commerce, social, and link-prediction domains.
On ogbn-arxiv, the largest graph, the GFT codebook-flip reaches attack success 0.75 and the
codebook do-operator still restores clean accuracy, so the causal localization holds at 169k
nodes. GraphAny breaks on PubMed and Amazon, and ZeroG breaks on the reddit and instagram social
graphs. Larger graphs dampen the structural edge attacks but not the feature, text, or codebook
carriers, which is consistent with the alignment layer, not the topology, being the reachable
surface.

\section{Classical-Attack Baselines}
\label{s:base}
A classical graph attack perturbs the adjacency, so it applies only to a model that reads an
adjacency, which restricts the comparison to OpenGraph node classification. We craft perturbed
adjacencies with DeepRobust (Metattack Meta-Self and PGD topology attack) against a two-layer GCN
surrogate on the model's own citation subgraph, editing only edges between real nodes and leaving
the class-prototype nodes untouched, then feed the same perturbed adjacency into OpenGraph and
score its real zero-shot \texttt{predict()} as a 10-pass average. Table~\ref{tab:base} shows that
these classical attacks are a strong baseline on this surface and exceed our tokenizer-targeted
attack, which we report rather than hide. There is no classical counterpart for the feature, text,
or codebook attacks that break the other five models, so those carriers have no baseline of this
kind.

\begin{table}[t]\centering\small
\setlength{\tabcolsep}{3pt}
\begin{tabular}{llrrrr}
\toprule
Dataset & Budget & Ours (spectral) & Metattack & PGD & Random \\
\midrule
Citeseer & 0.05 & 0.030 & 0.090 & 0.052 & 0.057 \\
Citeseer & 0.1 & 0.065 & 0.176 & 0.128 & 0.079 \\
Citeseer & 0.2 & 0.171 & 0.260 & 0.232 & 0.104 \\
Cora & 0.05 & 0.015 & 0.088 & 0.092 & 0.017 \\
Cora & 0.1 & 0.021 & 0.159 & 0.125 & 0.053 \\
Cora & 0.2 & 0.037 & 0.231 & 0.247 & 0.086 \\
PubMed & 0.05 & -0.059 & -- & 0.254 & 0.029 \\
PubMed & 0.1 & -0.006 & -- & 0.483 & -0.008 \\
PubMed & 0.2 & 0.304 & -- & 0.632 & 0.059 \\
\bottomrule
\end{tabular}

\caption{Baseline comparison on OpenGraph. ASR for our spectral-edge attack vs Metattack, PGD,
and random at matched edge budgets.}
\label{tab:base}
\end{table}

\section{Realizable Input-Space Attack on the Text Models}
\label{s:inputspace}
The threat model perturbs inputs, so for the text models the realizable attack edits the actual
class-description words, not the node or class embedding. The class-text attack is a joint
multi-class greedy HotFlip. For each class we tokenize its description, back-propagate the test
cross-entropy to the input-token embeddings, score every vocabulary swap by a first-order estimate,
re-encode the top candidates through the real sentence encoder, and accept the single best swap that
raises the loss subject to a cosine-similarity floor of $0.70$ to the original text, capped at eight
edits per class. Accuracy is the real decoder at every step, re-verified by injecting the edited
texts through the pipeline (max logit deviation $<2\times10^{-7}$). Table~\ref{tab:inputspace} gives
the realizable input-space attack success beside the representation-space upper bound, the node or
class embedding perturbation the attacker cannot emit. On citation graphs the text edit recovers
most of the upper bound. On the two social graphs it is exactly zero, and this is not a search
failure. The unconstrained class-embedding attack itself reaches only $0.02$ and $0.003$ there, so a
class-text carrier cannot move a two-class social decoder, and the $1.00$ in the upper-bound column
is a node-embedding perturbation with no realizable text form.

\begin{table}[t]\centering\small
\setlength{\tabcolsep}{4pt}
\begin{tabular}{llrr}
\toprule
Model & Dataset & Input-space & Rep.\ upper bound \\
\midrule
ZeroG & Cora      & 0.32 & 0.78 \\
      & Citeseer  & 0.69 & 1.00 \\
      & PubMed    & 0.57 & 1.00 \\
      & instagram & 0.00 & 1.00 \\
      & reddit    & 0.00 & 1.00 \\
OFA   & Cora      & 0.54 & 0.96 \\
      & PubMed    & 0.63 & 1.00 \\
\bottomrule
\end{tabular}
\caption{Realizable input-space class-text attack success versus the representation-space upper
bound for the text models. The upper bound is a node or class embedding perturbation the stated
attacker cannot produce. Input-space attack success is near the bound on citation graphs and zero on
the two social graphs.}
\label{tab:inputspace}
\end{table}

\section{Alignment-Targeted vs Generic Attack, and Black-Box Queries}
\label{s:e2e}
We hold the optimizer, the objective, and the budget fixed and vary only the search space. The
generic end-to-end attack optimizes one universal decision-space vector in the full embedding. The
alignment-targeted attack confines the same vector to the top sixteen principal directions of the
shared representation, the subspace used by align-then-ablate. Both are crafted on Cora and applied
unchanged to Cora, near-domain Citeseer, and far-domain PubMed at a matched budget
(Table~\ref{tab:e2e}). The generic vector is stronger on the source, because about half its budget
lies outside the shared subspace and overfits the source class geometry, and that part does not
transfer. On far-domain PubMed the generic vector reaches $0.065$, at the matched random control,
while only the alignment-targeted vector transfers above chance, to $0.224\pm0.002$ over three
seeds. This is what the alignment view buys, not a new optimizer but a perturbation that survives a
domain change. Near-domain Citeseer is fragile at
$0.27$ clean accuracy, so both vectors flip it and the two are equal there.

\begin{table}[t]\centering\small
\setlength{\tabcolsep}{4pt}
\begin{tabular}{llrrr}
\toprule
Budget ($\ell_2$) & Search space & Cora & Citeseer & PubMed \\
\midrule
1.34 & generic (full) & 0.563 & 0.344 & 0.022 \\
1.34 & alignment-targeted & 0.335 & 0.341 & 0.080 \\
2.68 & generic (full) & 0.762 & 0.344 & 0.060 \\
2.68 & alignment-targeted & 0.511 & 0.344 & 0.222 \\
\bottomrule
\end{tabular}
\caption{Generic end-to-end vs alignment-targeted attack, one universal vector crafted on Cora at
matched budget, attack success on each target. The generic vector wins on the source, the
alignment-targeted vector transfers to far-domain PubMed.}
\label{tab:e2e}
\end{table}

A black-box variant reads only the model's output scores, with no gradients. Score-based search on
ZeroG's text carrier reaches $73$ to $83$ percent of the white-box attack success at a few hundred
queries per node (Table~\ref{tab:blackbox}), so the threat model is evidenced at white box, gray
box (the cross-model spectral transfer above), and black box. This black-box result is at the
embedding carrier under query access, not raw-text queries.

\begin{table}[t]\centering\small
\setlength{\tabcolsep}{4pt}
\begin{tabular}{lrrr}
\toprule
Budget frac. & White-box & Black-box & Mean queries \\
\midrule
0.1 & 0.203 & 0.148 & 343 \\
0.2 & 0.385 & 0.307 & 305 \\
0.4 & 0.645 & 0.534 & 237 \\
\bottomrule
\end{tabular}
\caption{Score-based black-box attack on ZeroG's node-embedding carrier versus the white-box
attack, at matched budget.}
\label{tab:blackbox}
\end{table}

\section{Cross-Domain Transfer}
\label{s:transfer}
A single perturbation is crafted on Cora in the shared text space and applied verbatim to target
domains. Within the citation family it transfers (ZeroG Cora$\to$Citeseer 0.72, OFA
Cora$\to$PubMed 0.43, vs random 0.03). Across the citation-to-social gap it fails (ZeroG
Cora$\to$reddit 0.05 $\approx$ random 0.055, $\to$instagram 0.20). A naive universal additive
perturbation does not transfer at all. Only a generic semantic axis (top principal direction)
travels, and only through decoders that do not standardize per domain (ZeroG standardizes and
resists, OFA does not and is exposed). Cross-domain transfer rate (mean target ASR, excluding
source): OFA pcaAxis 0.43 (hit rate 1.0), ZeroG pcaAxis 0.36 (near/far split: Citeseer 0.72,
PubMed 0.01).

Matching the target class prototype in the shared space, the analog of an adversarial illusion
\citep{zhang2024illusions}, is the most budget-efficient within-family carrier. Crafted on Cora it
reaches Citeseer at attack success $0.65$ at a quarter of the budget where the task-loss vector
reaches only $0.11$, because the prototype direction is shared across the family while the task-loss
direction overfits the source geometry \citep{gupta2025advtransfer}. It still does not reach the far
domain, where no carrier beats a matched random baseline. This is the same split at a finer grain,
the shared direction travels within the family and no direction crosses to it.

\paragraph{Gray-box cross-model transfer (E5).} The attacker has no gradient access to the
victim. It crafts an edge set on a surrogate spectral model, using the surrogate's own white-box
attack, and applies the same edges verbatim to the victim, whose pipeline re-runs the singular
value decomposition, re-normalizes the adjacency, and, for AnyGraph, re-routes its experts. The
transfer is real but weak (Table~\ref{tab:crossmodel}): the victim drop exceeds the matched
flip-count random control at the larger budget, most clearly for OpenGraph into AnyGraph on
Citeseer, yet stays small in absolute terms. This is consistent with the spectral pair being the
resistant carrier, where even the white-box attack has little room on these high-accuracy tasks.
The mechanism, not white-box access, is what carries the effect, but the size of the effect is
bounded by the victim task's headroom.

\begin{table}[t]\centering\small
\setlength{\tabcolsep}{3pt}
\begin{tabular}{llrrrr}
\toprule
Direction & Data & Budget & Surrog. & Victim & Random \\
\midrule
Any$\to$Open & Citeseer & 10\% & 0.137 & 0.044 & 0.034 \\
             & Citeseer & 20\% & 0.207 & 0.095 & 0.075 \\
             & Cora     & 10\% & 0.248 & 0.028 & 0.029 \\
             & Cora     & 20\% & 0.343 & 0.086 & 0.058 \\
Open$\to$Any & Citeseer & 10\% & 0.062 & 0.014 & $-.00$ \\
             & Citeseer & 20\% & 0.159 & 0.095 & 0.001 \\
             & Cora     & 10\% & 0.020 & 0.015 & $-.00$ \\
             & Cora     & 20\% & 0.045 & 0.049 & 0.008 \\
\bottomrule
\end{tabular}
\caption{Gray-box cross-model transfer between the two spectral models, attack success (metric
drop), two seeds. Surrogate is the attacker's own model, victim is the target with no gradient
access, random is a matched flip-count edge control on the victim.}
\label{tab:crossmodel}
\end{table}

\section{Ablation Battery}
\label{s:abl}
\paragraph{A1: attack objective.} On a 4-block SBM through the reference tokenizer, task-loss and
fixed-basis-surrogate objectives both reach ASR 0.925, while subspace-rotation reaches 0.59
despite the largest misalignment. A gradient probe at the clean graph: subspace-objective
gradient norm $2.4\times10^{-6}$ vs task-loss $0.13$. The clean graph is a minimum of subspace
misalignment, so rotation is second-order there. We maximize task loss instead.

\paragraph{A3: rank dose-response.} Ablating rank $1..12$ of the alignment subspace on GFT gives
a graded collapse of the mediated fraction ($0.12\to0.32\to0.71\to0.74$). GraphAny's curve is
noisy because its phi-pathway effect is tiny (17/1000 nodes), consistent with the feature attack
routing through the channel logits rather than the distance features.

\paragraph{A5: spectral degeneracy.} Within a single graph the singular-value gap correlates with
misalignment absorbed per unit budget at $-0.69$: near-degenerate boundaries absorb 5--15$\times$
more rotation per edge flipped. The naive cross-graph proxy is confounded (correlation $+0.03$),
reported as an honest negative.

\paragraph{A9: targeted vs untargeted.} On GFT the targeted codebook-flip reaches higher ASR than
untargeted (0.71 vs 0.51) while flipping \emph{fewer} tokens (88.9\% vs 100\%): coordinated
redirection to a chosen token does more damage than scattering assignments.

\paragraph{A4: ablation method.} We compare four operators on the same alignment subspace:
plain projection, LEACE closed-form erasure, INLP, and a matched-rank random-subspace placebo.
On GFT's continuous embedding, projection removes the effect (mediated 0.77) but destroys clean
accuracy (clean-ablated 0.15). LEACE reduces the confound (mediated 0.43) yet still costs clean
accuracy (0.29). INLP preserves clean accuracy (0.78) but no longer removes the effect (mediated
0.06). No linear erasure both keeps clean accuracy and removes the attack, so the continuous
carrier does not localize. The discrete do-operator does. On GraphAny's distance features every
operator, including the random placebo, mediates 0.65 to 0.76 at no clean-accuracy cost, so a
rank-6 subspace of the 20-dimensional space is not specific. Both cases confirm that the clean
causal evidence is the discrete codebook, not a continuous subspace.

\paragraph{Additional surfaces and scale.} The attack also reaches auxiliary parts of the alignment
layer. Node injection perturbs the same tokenizer without editing existing edges, its effect growing
with the injected count, and AnyGraph's router can be sent to a close-substitute expert. Combining
input channels does not beat the strongest single one: a joint edge-and-feature attack on AnyGraph
and a joint text-and-edge attack on ZeroG each equal their best single lever, so the channels are
redundant, reaching the same representation rather than adding independent damage.

\section{Robust GFM: Full Tables}
\label{s:rgfm}
\paragraph{C2 certified robustness (real GFT).} Randomized smoothing on the representation,
certified via Cohen et al.\ (2019). Radii are in carrier space (input-space requires $\Phi$
Lipschitz).

\begin{table}[t]\centering\small
\begin{tabular}{lrrr}
\toprule
$\sigma$ & \% certified & certified acc & median radius \\
\midrule
0.10 & 96.5 & 0.772 & 0.270 \\
0.25 & 88.6 & 0.739 & 0.577 \\
0.50 & 73.2 & 0.651 & 0.630 \\
\bottomrule
\end{tabular}
\caption{Certified robustness of GFT under representation-space smoothing.}
\label{tab:cert}
\end{table}

\paragraph{C1 joint encoder+codebook retrain (real GFT, Cora).} Table~\ref{tab:jr}: the joint
retrain moves the geometry offline hardening could not (margin, mean flip budget, flip rate), but
end-to-end ASR gets worse, because separating tokens makes flips rarer yet each surviving flip
more damaging on the fused readout.

\begin{table}[t]\centering\small
\setlength{\tabcolsep}{4pt}
\begin{tabular}{lrrr}
\toprule
metric & vanilla & R-GFT & $\Delta$ \\
\midrule
clean acc & 0.782 & 0.781 & $-0.002$ \\
mean margin & 0.521 & 0.771 & $+0.250$ \\
flip budget (median $\ell_2$) & 1.380 & 1.383 & $+0.003$ \\
flip budget (mean $\ell_2$) & 1.456 & 2.044 & $+0.588$ \\
ASR @$\ell_2{=}4$ & 0.461 & 0.589 & $+0.128$ \\
flip\% @$\ell_2{=}4$ & 100.0 & 98.8 & $-1.2$ \\
\bottomrule
\end{tabular}
\caption{Joint-retrain R-GFT vs vanilla GFT. The margin objective is counterproductive on a
codebook: rarer but more potent flips.}
\label{tab:jr}
\end{table}

\paragraph{Static defense (attack success, defended vs undefended).} Against a defender-unaware
attacker, GraphAny standardization gives $0.377\to0.107$ at a 1.1 point clean cost and smoothing
($\sigma{=}0.5$) gives $0.377\to0.187$ at no cost. On GFT's codebook both fail: smoothing rises
from $0.501$ to $0.824$ as $\sigma$ grows, and standardization gives $0.501\to0.530$. Simple
training-free defenses from prior work recover only about ten points, and some (low-rank
projection, embedding smoothing) backfire.

\paragraph{Adaptive defense (the defenses do not hold).} We re-run the GraphAny defenses against
an attacker that knows them. Standardization is a differentiable map, so an attacker that
differentiates through it recovers the full attack: static $0.107$ becomes $0.391$ under the
adaptive attack, at or above the undefended $0.377$. For smoothing, an attacker that averages its
gradient over the noise (expectation over transformations, 32 samples) recovers most of the
benefit: the static $0.187$ at $\sigma{=}0.5$ becomes $0.305$, and at lower $\sigma$ the adaptive
attacker recovers essentially all of it. Margin hardening on the codebook already raises attack
success non-adaptively, so no adaptive attack is needed. None of the training-free defenses
survives an adaptive attacker.

\paragraph{GFT codebook defenses, static and adaptive (three seeds).} On the discrete codebook the
picture is worse than on the continuous carrier. The direct defenses do not lower attack success
even against a static attacker, because the noise they add flips tokens, and an adaptive attacker
raises it further (Table~\ref{tab:gftadapt}). There is no static gain to recover.

\begin{table}[t]\centering\small
\setlength{\tabcolsep}{4pt}
\begin{tabular}{lrrrr}
\toprule
Defense & Clean acc & Undef. & Static & Adapt. \\
\midrule
standardize & 0.756 & 0.493 & 0.513 & 0.666 \\
smoothing $\sigma{=}0.25$ & 0.784 & 0.493 & 0.675 & 0.722 \\
smoothing $\sigma{=}0.5$ & 0.783 & 0.493 & 0.809 & 0.868 \\
\bottomrule
\end{tabular}
\caption{GFT codebook defenses at budget four. The defenses raise attack success even statically,
because the added noise flips tokens, and the adaptive attacker raises it further. Standard
deviation across seeds is $0.001$ to $0.003$.}
\label{tab:gftadapt}
\end{table}

\paragraph{Token-flip-insensitive decoder (proof of concept).} We change only the decoder to
down-weight the flippable code head, keeping the encoder, the codebook, and the attack fixed
(Table~\ref{tab:decoder}). Down-weighting cuts the codebook-flip attack from $0.493$ to $0.317$ at
budget four for a fraction of a point of clean accuracy, and to $0.146$ in the proto-only limit at
about one point. At budget one the proto-only decoder cuts a single flip from $0.102$ to $0.018$.
It defeats the token flip specifically. An adaptive attacker moves onto the continuous prototype
head, where the number matches the vanilla decoder, so the restriction does not make the layer
robust to an unrestricted perturbation. The soft top-$m$ readout helps only under transfer and adds
an easier continuous surface that an adaptive attacker exploits.

\begin{table}[t]\centering\small
\setlength{\tabcolsep}{3pt}
\begin{tabular}{llrrr}
\toprule
Budget & Decoder & Clean & Transfer & Adaptive \\
\midrule
1 & vanilla & 0.785 & 0.102 & 0.231 \\
1 & proto-only & 0.773 & 0.018 & 0.235 \\
4 & vanilla & 0.785 & 0.493 & 0.766 \\
4 & down-wt $t{=}.05$ & 0.784 & 0.317 & 0.772 \\
4 & proto-only & 0.773 & 0.146 & 0.766 \\
\bottomrule
\end{tabular}
\caption{Token-flip-insensitive decoder against the codebook-flip attack. Down-weighting the code
head reduces the transfer attack at small clean cost. The adaptive column shows the attacker moving
onto the prototype head.}\label{tab:decoder}
\end{table}

\section{Method, Datasets, and Reproduction}\label{s:method}
\paragraph{Threat model detail.} \emph{Capability} per carrier: edges and the features that enter
the SVD for the spectral models; input node features for GraphAny; node text or the shared
class-description text for the text models; and the pre-quantization embedding for the codebook,
which has no realized input form. Every perturbation is bounded, stated as a fraction of the edges
or injected nodes, an $\ell_\infty$ (per-entry) or $\ell_0$ (number-of-entries) bound on features,
or a cosine-similarity floor on text. Two fidelity caveats bound the efficacy numbers without
changing which layer is attacked: OFA's trained RGCN decoder is not run, so its efficacy uses a
text-space stand-in, and ZeroG is evaluated in its baseline pipeline with LoRA and belief
propagation off.

\paragraph{Spectral tokenizer.} Following OpenGraph: $\bar A = D^{-1/2}AD^{-1/2}$;
$U,S,V=\mathrm{SVD}(\bar A)$, node features $\mathrm{LN}(U\sqrt S + V\sqrt S)$ smoothed by
$\sum_{l=1}^{L}\bar A^{l}$. AnyGraph adds a feature SVD and a top-1 self-scoring mixture-of-experts
router. GraphAny solves closed-form LinearGNN channels and fuses them with an entropy-normalized
attention. The spectral attack optimizes edge flips in a continuous relaxation projected to an
$\ell_0$ budget. The fixed-basis surrogate detaches the singular basis and differentiates the
smoothing operator, avoiding SVD-backpropagation instability at near-degenerate spectra.

\paragraph{Codebook.} GFT uses a cosine, 4-head codebook ($C\in\mathbb{R}^{4\times128\times768}$);
assignment is $\arg\max_c \cos(\mathrm{proj}_h(z), C[h,c])$ per head. The token-flip attack
minimizes the boundary margin toward the second-nearest token with a straight-through estimator.
The do-operator pins the discrete indices to their clean values.

\paragraph{Align-then-ablate.} We identify the alignment subspace by fitting an orthogonal
Procrustes map between paired representations of the same content in two domains, then taking the
rank $k$ at which the cross-domain CKA between the aligned representations peaks, and $P$ is the span
of the aligned top-$k$ directions. We ablate it by projection $I-PP^\top$ and, for the continuous
carriers, also with closed-form linear erasure (LEACE) and iterated nullspace projection (INLP).
We report the mediated fraction $1-(A^{\rm abl}_{\rm clean}-A^{\rm abl}_{\rm atk})/(A_{\rm
clean}-A_{\rm atk})$ against a matched random-subspace placebo, with a dose-response over rank.
The discrete codebook do-operator is the only intervention that localizes cleanly. No linear
erasure on a continuous carrier both preserves clean accuracy and removes the attack.

\paragraph{Metric and hyperparameters.} Attack success is the relative accuracy degradation
$\mathrm{ASR}=(A_{\rm clean}-A_{\rm atk})/A_{\rm clean}$, where $A$ is node-classification accuracy,
or Recall@20 for AnyGraph's link-prediction task. All attacks are untargeted unless stated. The
feature and codebook attacks use projected gradient ascent, $T{=}60$ steps for features and
$T{=}150$ for edges, step size $\eta{=}\epsilon/4$ under an $\ell_\infty$ budget and top-$k$
projection under an $\ell_0$ or edge budget. Budget sweeps: edge flips at $5,10,20\%$ of the real
edges, feature $\ell_\infty\in\{0.05,0.1,0.2\}$ and $\ell_0\in\{5,10,25,50\}$ entries per node, and
codebook margin budget $\in\{1,2,4\}$. Each cell averages over three seeds unless a single seed is
noted. The attacks are cheap. The feature and codebook attacks run in seconds to a couple of minutes
on one CPU or GPU, and the edge attack in a few minutes at 150 gradient steps, so crafting one
attacked input costs on the order of the model's own inference pass over the dataset.

\paragraph{Models and datasets.} Six public GFMs (OpenGraph, AnyGraph, GraphAny, OFA, GFT, ZeroG),
each reproduced to its published clean number before attacking and wrapped behind one interface.
Datasets span citation (Cora, Citeseer, PubMed, ogbn-arxiv), web (WikiCS), e-commerce
(Amazon-Photo/Computers, ogbn-products), heterophily (Cornell, Texas, Wisconsin), and social
(reddit, instagram). Attacks are evaluated on each model's real inference pipeline. White-box
gradients that pass through a non-differentiable step (e.g.\ the entropy-normalized distance in
GraphAny) use a torch surrogate validated to $\sim$$10^{-6}$, with every number re-evaluated on
the real pipeline. Code and configuration will be released.

\section{Headroom Law: Held-Out Prediction}
\label{s:headroom}
\begin{figure}[t]
\centering
\includegraphics[width=\columnwidth]{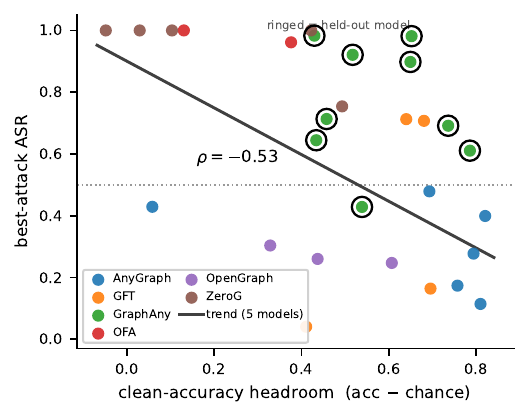}
\caption{Reachability against clean-accuracy headroom across twenty-nine cells and six models.
Each point is one model on one dataset. Headroom is clean accuracy above chance and reachability
is the strongest attack in the cell. The trend is fit on five models and the held-out model is
ringed. Headroom orders the cells and the attack carrier sets the vertical offset.}
\label{fig:headroom}
\end{figure}

We turn the headroom observation into a held-out prediction. A cell is one model on one dataset.
Headroom is reproduced clean accuracy minus chance, where chance is one over the class count for
node classification and zero for AnyGraph's Recall@20 link task. Reachability is the largest
mean-over-seed attack success any attack and budget reaches in the cell, and a cell is reachable
when that exceeds one half. Over twenty-nine cells the Spearman correlation between headroom and
best attack success is $-0.53$ (95\% bootstrap CI $[-0.80,-0.16]$ over ten thousand cell
resamples, $99.7\%$ negative, $p=0.003$), stable under three summaries of attack success and
stronger with the chance subtraction than without it. Scoring reachability by negative headroom
gives an AUC of $0.68$ (Figure~\ref{fig:headroom}). For a held-out test we leave out one whole
model, fit a line on the other five, and predict the held-out model. Every fold returns a
negative slope between $-0.36$ and $-0.76$, so the direction of the law does not depend on any
single model, and the held-out reachable AUC is $0.68$. Out-of-sample point prediction is weaker:
the pooled rank correlation between predicted and true attack success is $0.22$ and the mean
absolute error, $0.31$, does not beat predicting the global mean, $0.28$. Leaving out a whole
domain family instead of a model does slightly better, with a pooled predicted-versus-true
Spearman of $0.45$ ($p=0.014$) and mean absolute error $0.26$ just under the mean baseline. The
residual variance tracks the model, not the domain, which is what we expect if the model's
alignment carrier sets the offset while headroom sets the order. We estimate the carrier gain
$\kappa$ per cell as the through-origin initial slope of the attack's success-versus-budget curve,
in that model's budget unit. Under the same model holdout, adding $\kappa$ to headroom lowers
held-out error from $0.29$ to $0.20$ and raises the predicted-versus-true rank correlation from
$0.32$ to $0.62$. Because $\kappa$ is read from the same curve whose peak is the target, we rerun
with $\kappa$ taken from low budgets only and the top-budget point dropped, on the cells with at
least three budgets: the two-factor fit still wins, error $0.29$ to $0.26$ and rank correlation
$0.49$ to $0.65$. The literal ratio does not transfer across models, since $\kappa$ is in
model-specific units, so we present $\kappa$ as a measured mechanistic factor, not a cross-model
formula. The headroom law is a robust ordering rule and a suggestive, not conclusive, quantitative
predictor.

\paragraph{The headroom law does not survive on realizable attacks.} The pooled $-0.53$ was computed
on a set that mixed realizable and representation-space attacks, with the unrealizable
node-embedding cells at attack success one sitting at low headroom. Restricting to the realizable
input-space attacks, GraphAny features, spectral edges, and the text-model class-text edits, and
excluding the codebook cells and the $h=0$ link cells, the correlation vanishes and flips sign:
Spearman $+0.01$ on $N=21$ cells ($p=0.98$, CI $[-0.50,+0.48]$), or $+0.02$ with the link cells. A
permutation test on the within-model ranks, which is adequately powered where per-model tests on
three datasets are not, is not significant (pooled within-model statistic $-0.23$, $p=0.25$
one-sided, ten thousand permutations). The per-model correlations are mixed, negative for OpenGraph
and GraphAny and positive for AnyGraph and ZeroG. Controlling for the carrier gain the partial
Spearman is $+0.15$ with a confidence interval crossing zero. What ranks with attack success is the
carrier gain. On realizable attacks $\kappa$ and attack success correlate at $+0.72$ across cells
($p=0.0003$) and $+0.90$ across the six models, while headroom is flat ($+0.20$). We caution that
$\kappa$ is read from the attack curve, so its correlation with peak success is partly a
re-description, and its units differ across carriers, so it is a comparative descriptor and not a
mechanism. The honest state is that clean accuracy headroom does not predict reachability on
realizable attacks, and the carrier gain is the better descriptor.

\section{Representation-Space Upper Bound on the Spectral Models}
\label{s:repub}
To separate the input lever from the alignment layer, we perturb the representation the decoder
reads directly, at a per-node $L_2$ budget swept to saturation, and report the peak over three
seeds against a matched-norm random control (Table~\ref{tab:repub}). For OpenGraph we perturb the
spectral token $E$ the transformer reads, for AnyGraph the expert-input token, for GraphAny the
pre-fusion channel logits. The upper bound is high for all three, and the matched random control
stays below $0.03$, so the effect is the crafted direction. The decisive cell is OpenGraph on Cora,
the dataset that resists the edge attack: a direct token perturbation collapses accuracy from
$0.75$ to $0.01$. OpenGraph and AnyGraph are therefore fragile at the alignment layer, and the edge
channel is a weak lever on it, which the small carrier gain records.

\begin{table}[t]\centering\small
\setlength{\tabcolsep}{4pt}
\resizebox{\columnwidth}{!}{%
\begin{tabular}{llrrr}
\toprule
Model & Rep.\ perturbed & Rep.\ UB & Random & Input peak \\
\midrule
GraphAny & channel logits & 0.99 & 0.27 & 0.98 \\
OpenGraph & spectral token $E$ & 0.99 & 0.03 & 0.30 \\
AnyGraph & expert-input token & 0.82 & 0.02 & 0.47 \\
\bottomrule
\end{tabular}}
\caption{Representation-space upper bound for the spectral models, perturbing the decoder-input
representation directly. High for all three, so the alignment layer is fragile even where the
input-space edge attack is weak.}
\label{tab:repub}
\end{table}

\paragraph{Budget commensurability and on-manifold distance.} The rep-space budget is a per-node
$\ell_2$ radius reported as a fraction of the mean representation norm. Collapsing a model needs a
fraction of $0.2$ to $1.6$, so a perturbation worth $20$ to $160$ percent of the representation's
own norm, while a matched-norm random perturbation never exceeds $0.03$ attack success, so the
direction carries it. This is not commensurable with the input budget. OpenGraph's fall from $0.75$
to $0.01$ needs a token perturbation of $0.4$ to $0.8$ of the token norm, whereas an edge edit at a
fifth of the edges moves the token below the $0.05$ threshold, because the singular value
decomposition and the power-sum smoothing attenuate edge edits. So the rep-space number is a
large-budget upper bound on fragility, not a small-perturbation vulnerability. Whether the attack
leaves the clean manifold depends on carrier dimension. GraphAny's $21$-dimensional channel-logit
carrier moves off manifold and is caught at AUC $1.0$ exactly when it becomes effective, while the
high-dimensional spectral tokens of OpenGraph ($1024$) and AnyGraph ($512$) stay largely on
manifold even at collapse (best detector AUC $0.72$ and $0.67$), because the effective direction
lies inside the high-variance clean subspace.

\paragraph{Plain-network control.} A perturbation of order the representation norm collapses any
classifier, so we ran the control. We trained a plain single-domain GCN, GraphSAGE, and GAT on Cora,
Citeseer, and PubMed and applied the identical directed attack to their penultimate node embedding.
All nine collapse at a fraction $0.48$ to $0.70$ of the representation norm, mean $0.61$, with a
matched-norm random control below $0.06$ (Table~\ref{tab:gnnctrl}). GraphAny ($0.62$) and AnyGraph
($0.56$) sit in this band, so their representation-space collapse is the generic fragility of a
message-passing network. OpenGraph collapses at $0.12$, about five times below the plain-network
band, the one alignment-specific case. The collapse threshold correlates with carrier dimension
across the three spectral models (Pearson $-0.88$, $p=0.02$), but with the sign opposite to
Claim~3, higher-dimensional carriers collapsing at a lower budget, so the collapse axis does not
support fragility-from-concentration. This is a consistency check on six points, not a law.

\begin{table}[t]\centering\small
\setlength{\tabcolsep}{4pt}
\resizebox{\columnwidth}{!}{%
\begin{tabular}{llr}
\toprule
Model & Carrier (dim) & Collapse frac \\
\midrule
plain GNN (GCN/SAGE/GAT) & penultimate & 0.61 \\
OpenGraph & spectral tok.\ (1024) & \textbf{0.12} \\
AnyGraph & joint-SVD tok.\ (512) & 0.56 \\
GraphAny & channel logits (21) & 0.62 \\
\bottomrule
\end{tabular}}
\caption{Representation-space collapse threshold, the fraction of the representation norm at which a
directed attack drives attack success past one half. Only OpenGraph is below the plain-network band.}
\label{tab:gnnctrl}
\end{table}

\paragraph{No second alignment-specific case.} We also test the non-spectral carriers against the
tighter same-representation control, a plain classifier trained on the very representation the
decoder reads. None is more fragile than its own baseline. The GFT codebook decoder collapses at
frac $0.55$ against $0.60$ for a linear probe on the same embedding, since its prediction is
dominated by a continuous nearest-prototype head. The text decoders of ZeroG and OFA collapse at a
low frac ($0.14$), but so does a plain probe on the same sentence embedding ($0.13$ for OFA, lower
for ZeroG), so the low threshold is a property of the sentence-embedding representation, not the
alignment decoder, and ZeroG's cosine read-out is in fact more robust than the probe because it
normalizes to the sphere. So OpenGraph remains the single carrier whose alignment layer is fragile
beyond a matched plain classifier, and we report it as one case, not a class property.

\paragraph{The OpenGraph case is the tokenizer, not the decoder.} The same-representation control is
sharpest on OpenGraph itself. We read its clean spectral token with two plain classifiers, a linear
softmax probe and a two-layer perceptron, both trained on the clean tokens of non-test nodes and hit
with the identical directed attack. On Cora the linear probe collapses at frac $0.04$ and the
perceptron at $0.09$, against the transformer decoder's $0.15$. On Citeseer the three are $0.03$,
$0.05$, and $0.09$. On the identical token both probes collapse at or below the decoder, so the
decoder is the most robust of the three readers and does not amplify the push. The low threshold is a
property of the singular value decomposition token, which sits at a large radius with a small logit
margin, and any classifier on it inherits the fragility. The matched-norm random control is
near-inert. Two caveats do not reverse the direction. The probe attack is deterministic while the
decoder attack must be robust to the transformer's random anchor sampling, so part of the decoder's
higher threshold may be attack difficulty rather than robustness, and the probes are fit to the
target graph while the decoder is zero-shot. In both the decoder is at most as fragile as its token,
never more, so the reading that the trained transformer manufactures the fragility is excluded. Data
in \texttt{results/opengraph\_probe.json}.

\begin{figure}[t]
\centering
\includegraphics[width=0.62\columnwidth]{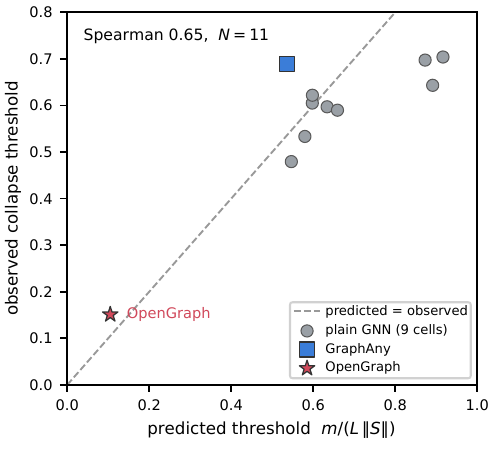}
\caption{Predicted against observed representation-space collapse threshold. The predicted value is
the structural, attack-curve-free flip budget $m/(L\,\|S\|)$ from the decoder's local Lipschitz
constant and the clean margin and representation norm, with no attack curve. Nine plain-network cells
and GraphAny fall near the diagonal, and OpenGraph is the extreme low point, its threshold predicted
from clean representations alone.}
\label{fig:predobs}
\end{figure}

\paragraph{A structural, attack-curve-free carrier gain.} The fitted $\kappa$ is read from the attack
curve and is not comparable across carriers. A decoder-side structural form is. We take the local
Lipschitz constant $L=\|\partial g/\partial S\|_2$ of each decoder, the top singular value of its
Jacobian at clean test representations, by autograd power iteration. The flip-budget proposition then
predicts a per-node collapse threshold $m/(L\,\|S\|)$, a margin of logit change divided by the
decoder sensitivity and the clean representation norm. This structural prediction tracks the measured
collapse threshold across the classification decoders and the plain networks
(Figure~\ref{fig:predobs}, Spearman $0.65$, $N=11$, mean absolute error $0.10$ in the frac unit) and
reproduces OpenGraph's $0.12$ from clean representations alone. OpenGraph sits at the extreme low
corner, a predicted and an observed threshold both far below the plain-network cluster on the
diagonal. Nine of the eleven cells are plain networks and OpenGraph is a leverage point, so the
correlation is not significant with OpenGraph held out (Spearman $0.53$, $N=10$, $p=0.12$), and only
two cells are graph foundation models. We therefore report $m/(L\,\|S\|)$ as a rank-ordering
heuristic for where a carrier sits, not a quantitative law. It is dimensionless, so it is invariant to the per-carrier unit that made the
fitted $\kappa$ swing under rescaling, which is why it succeeds where the input-side Davis-Kahan gap
was degenerate. The mechanism is representation geometry, not decoder amplification. Although
dot-product self-attention is not globally Lipschitz \citep{kim2021lipschitz}, the per-node local
Lipschitz constant of OpenGraph's trained transformer is $0.81$, below a linear probe on the same
token ($15.5$) and below the plain-network head band ($1.5$ to $5.7$), so the LayerNorm leaves the
decoder near-isometric per node. OpenGraph is fragile because $L\,\|S\|$ is large through a large
token radius and a small margin, not because attention amplifies the push. Link-prediction AnyGraph
is excluded from the law because its decoder outputs candidate scores, not class logits, so $L$ is
not in comparable units. Data in \texttt{results/lipschitz\_kappa.json}.

\paragraph{Spectral rotation edge attack.} On the spectral tokenizer $E=(\sum_l \bar A^l)\,
\mathrm{LN}(U\sqrt\Sigma+V\sqrt\Sigma)$, projected gradient ascent on edges is inert. The singular
basis $U,V$ is detached from the gradient, since a differentiable full singular value decomposition
backward is numerically unstable at OpenGraph's near-degenerate spectrum, so the gradient carries
only the smoothing term and its edge attack sits at the noise floor, with a negative attack success
($-0.01$ to $-0.02$) that helps the model. We instead compute the per-edge score $\langle g_E,\,
dE/dA_{ab}\rangle$ analytically, where $g_E$ is the decoder gradient with respect to the token (one
backward pass, no decomposition) and $dE/dA_{ab}$ includes the eigenvector-rotation term the gradient
drops. For a symmetric $\bar A=Q\Lambda Q^\top$, flipping edge $(a,b)$ gives $d q_k=\sum_{m\neq k}
(q_m^\top d\bar A\,q_k)/(\lambda_k-\lambda_m)\,q_m$, and we clamp the near-degenerate denominators.
This rotation term is seven times the norm of the smoothing term. On a well-separated synthetic
spectrum the analytic gradient matches a finite-difference derivative to relative error $3\times
10^{-7}$ and converges as $h\to0$, so the perturbation theory is correct, and the token reconstruction
matches the real tokenizer to $1.4\times10^{-7}$. On OpenGraph's real $\bar A$ the median singular gap
is near $4\times10^{-4}$, the derivative is ill-defined (the finite-difference error does not
converge), and the low-rank solver returns a different basis on each draw with the decoder invariant
to that rotation. The regularized attack, evaluated on the real pipeline over three basis draws, still
moves OpenGraph off the noise floor, to attack success $0.15$ at a tenth of the edges and $0.16$ at a
fifth, against a matched random flip of $0.03$ and $0.09$ and the negative gradient attack. So
OpenGraph's edge resistance is a property of the degenerate spectrum, not intrinsic robustness, and
the same degeneracy caps the attack below the alignment and text carriers of the other models. Data
in \texttt{results/spectral\_rotation\_attack.json}.

\section{Defenses: Full Table and Detection}
\label{s:deffull}
Table~\ref{tab:defense} is the full defense evaluation under a static and an adaptive attacker.
None of the robustification defenses survives, on the codebook they backfire even statically, and
the off-the-shelf GCN-SVD baseline is unusable on the spectral carrier.

\begin{table}[t]\centering\small
\setlength{\tabcolsep}{3pt}
\begin{tabular}{lcccc}
\toprule
Carrier / defense & Clean$_{\text{def}}$ & Undef. & Static & Adaptive \\
\midrule
\multicolumn{5}{l}{\emph{Continuous, GraphAny, fusion-dist PGD, $L_2{=}2.0$}}\\
\quad standardize (C3)              & 0.783 & 0.377 & \textbf{0.107} & 0.391 \\
\quad smoothing $\sigma{=}0.5$ (C2) & 0.797 & 0.377 & 0.186 & 0.306 \\
\midrule
\multicolumn{5}{l}{\emph{Codebook, GFT, codebook-flip $L_2{=}4.0$}}\\
\quad standardize (C3)              & 0.756 & 0.493 & 0.513 & 0.666 \\
\quad smoothing $\sigma{=}0.25$ (C2)& 0.784 & 0.493 & 0.675 & 0.722 \\
\quad robust decoder (ours)         & 0.784 & 0.493 & \textbf{0.317} & 0.772 \\
\midrule
\multicolumn{5}{l}{\emph{Spectral, OpenGraph, edge 20\%, GCN-SVD $r{=}50$}}\\
\quad Citeseer                      & 0.417 & 0.146 & 0.063$^{\dagger}$ & -- \\
\quad PubMed                        & 0.399 & 0.170 & 0.364 & -- \\
\bottomrule
\end{tabular}
\caption{Attack success (lower is better) with and without each defense. Clean$_{\text{def}}$ is
accuracy under the defense, against undefended clean accuracy $0.794$, $0.785$, $0.607/0.667$.
Budgets differ by carrier, stated per block. The GraphAny fusion-dist budget is smaller than its
headline feature budget. GCN-SVD costs nineteen points of clean accuracy on Citeseer, so the
$^{\dagger}$ value is a clean-accuracy collapse, not a defense.}
\label{tab:defense}
\end{table}

\paragraph{Detection.} We fit a density test on clean representations of an evaluation split and
score held-out clean versus attacked representations, reporting AUC and the true-positive rate at a
five percent false-positive rate (Table~\ref{tab:detect}). On the low-dimensional continuous
carriers detection is near perfect and holds under an adaptive attacker that adds a
stay-on-manifold penalty, read as a two-sided density test since the penalty can overshoot into an
anomalously typical region. On GraphAny the adaptive two-sided AUC stays at $0.74$ while the evasion
drives attack success from $0.37$ to $0.01$, so evasion and attack are in tension. On GFT the
continuous pre-quantization embedding holds at $0.88$ even where the discrete codebook-native scores
are evadable. The high-dimensional text embedding of ZeroG is the weak case, where a low-budget
perturbation hides in a near-isotropic space.

\begin{table}[t]\centering\small
\setlength{\tabcolsep}{4pt}
\resizebox{\columnwidth}{!}{%
\begin{tabular}{llrrr}
\toprule
Model & Detector & AUC & TPR@5\% & Adapt.\ 2-sided \\
\midrule
GraphAny & Mahalanobis & 1.00 & 1.00 & 0.74 \\
GFT      & Mahalanobis-$z$ & 1.00 & 1.00 & 0.88 \\
ZeroG    & class Mahalanobis & 0.66 & 0.08 & 0.92 \\
\bottomrule
\end{tabular}}
\caption{Detection of attacked inputs by a density test fit on clean representations. Near perfect
on the low-dimensional continuous carriers and holding under an adaptive attacker, weak on the
high-dimensional text embedding.}
\label{tab:detect}
\end{table}

\paragraph{Adaptive detection pareto.} We sweep the stay-on-manifold penalty weight $\lambda$ and
report at each value both the attack success and the two-sided detection AUC (Table~\ref{tab:pareto},
two seeds). On GraphAny the two are traded off. Raising $\lambda$ drives attack success down and
lets the attacker approach chance detection, but the strongest evasion that reaches AUC $0.54$
already costs two thirds of the attack, and the full-strength attack is fully detected. On GFT the
two-sided test holds near AUC $0.87$ at every $\lambda$, so there is no penalty that keeps the
codebook attack strong and hides it. No setting on either carrier gives a strong attack that also
escapes the monitor.

\begin{table}[t]\centering\small
\setlength{\tabcolsep}{5pt}
\begin{tabular}{lrr|lrr}
\toprule
\multicolumn{3}{c|}{GraphAny} & \multicolumn{3}{c}{GFT} \\
$\lambda$ & ASR & AUC$_2$ & $\lambda$ & ASR & AUC$_2$ \\
\midrule
0.000 & 0.37 & 1.00 & 0.00 & 0.48 & 1.00 \\
0.005 & 0.12 & 0.54 & 0.05 & 0.58 & 0.87 \\
0.010 & 0.06 & 0.60 & 0.20 & 0.38 & 0.88 \\
0.020 & 0.03 & 0.72 & 1.00 & 0.23 & 0.88 \\
0.050 & 0.01 & 0.75 & 5.00 & 0.18 & 0.88 \\
\bottomrule
\end{tabular}
\caption{Adaptive detection pareto. Attack success and two-sided detection AUC as the
stay-on-manifold penalty $\lambda$ sweeps. GraphAny trades attack strength for evasion, while GFT's
two-sided detector does not drop.}
\label{tab:pareto}
\end{table}

\paragraph{A feature-distribution attack does not beat task loss on this surface.} A reader may ask
whether a stronger attack than projected gradient ascent on task loss exists for the alignment layer.
The natural candidate is the feature-distribution attack \citep{inkawhich2020transferable,inkawhich2019feature},
which in vision moves a representation into a target class's feature density rather than across the
decision boundary and transfers better than a task-loss attack. We adapt it to the alignment layer
and call it Aligned-FDA. We fit a class-conditional Gaussian in the shared subspace and push a node
toward a wrong-class density and away from its own, with a Mahalanobis objective in place of
cross-entropy. A pre-check confirms this is not a relabeling of targeted gradient ascent. The class
covariances in the subspace are anisotropic (condition number $17.6$ at rank $20$), the classes
overlap (linear separability $0.75$), and the Mahalanobis attack direction differs from the centroid
direction for most nodes. The attack is nonetheless weaker. In-domain on ZeroG Cora it reaches attack
success $0.23$, $0.56$, and $0.78$ across three budgets, against $0.75$, $0.97$, and $1.00$ for
task-loss ascent, because a class-conditional target is a harder objective than crossing the nearest
boundary. Its cross-domain transfer to Pubmed is no better than the task-loss universal vector or a
centroid push. This is consistent with the result that representation-space perturbations transfer
poorly because they target model-specific geometry rather than shared input structure
\citep{gupta2025advtransfer}, and a class-conditional direction is the most model-specific of the
three we try, while the plainer centroid direction transfers best. Against the detector it is the sharpest form of the no-free-evasion property, and the two carriers
tell it in two ways. On GraphAny the Aligned-FDA attack drives the class-conditional detector below
chance, since it makes the representation more typical of a wrong class than a clean input, but it
lands in a cluster the decoder does not read as that class, so its attack success is $0.015$ while a
task-loss attack at the same budget reaches $0.38$ and is fully detected. The codebook carrier is the
harder test, since GFT's prototype head reads the representation directly. There the attack does flip
the prediction, at attack success $0.66$ against $0.46$ for the task-loss attack, and it evades both
the class-conditional detector and the codebook-specific residual and margin tests, which fall to
$0.60$ and $0.54$. A global two-sided density test on the representation still flags it, at AUC
$0.85$ across budgets. So a class-conditional or codebook detector alone is not enough against a
feature-distribution attack, which is why we monitor with the global two-sided density test, and on
that test no attack both fools the model and evades the monitor. Data in
\texttt{results/aligned\_fda\_*.json}.


\section{Theory: Fragility, Headroom, and Absorption}
\label{s:theory}

We formalize the three empirical rules of Section 3. Throughout, the model is $g\circ\Phi$.
The alignment map $\Phi$ sends an input $x=(A,X,\text{text})$ to a representation $r\in S$.
For a spectral model $S$ is the leading rank-$k$ singular subspace of the normalized adjacency
$\bar A=D^{-1/2}AD^{-1/2}$, with singular values $\sigma_1\ge\sigma_2\ge\cdots$ and token
features $E=\big(\sum_{l=1}^{L}\bar A^{l}\big)\,\mathrm{LN}(U\sqrt\Sigma+V\sqrt\Sigma)$. The
graph is undirected, so $\bar A$ is symmetric and its SVD is its eigendecomposition with
$U=V$. The decoder $g$ reads $r$ and is $L$-Lipschitz on $S$. The attacker perturbs the input
by $\delta$ within a budget $B$, measured by $\|\delta\|$ for features and by the edit count
for edges. We write $P_k$ for the orthogonal projector onto the retained subspace and
$\Theta(P_k,P_k')$ for the principal angles between the clean and perturbed subspaces.

\subsection{P1: Fragility of a Low-Dimensional Shared Bottleneck}

\paragraph{Assumptions.}
\begin{itemize}
\item[(A1)] $\bar A$ is symmetric with a spectral gap $\gamma_k=\sigma_k-\sigma_{k+1}>0$ at the
boundary of the retained subspace.
\item[(A2)] The input-to-operator map is Lipschitz. A perturbation of budget $B$ produces a
symmetric $E=\bar A(x+\delta)-\bar A(x)$ with $\|E\|_F\le c_A B$ for a constant $c_A$ fixed by
the normalization.
\item[(A3)] The decoder reads the boundary direction. The runner-up singular direction
$u_{k+1}$ has nonzero image under $g$, with directional gain $\mu=\|\partial_{u_{k+1}} g\|>0$.
\end{itemize}

\paragraph{Proposition P1.}
\emph{Under (A1)-(A3) there is a perturbation of budget $B$ whose induced output change obeys}
\[
\big\|g(\Phi(x{+}\delta))-g(\Phi(x))\big\|\;\ge\;\mu\,c\,\sigma_k\,\frac{B}{\gamma_k}\;+\;o(B),
\]
\emph{for a constant $c>0$, and every perturbation of budget $B$ obeys the matching upper bound
$L\,c_A'\,\sigma_1\,B/\gamma_k$. The amplification factor is $1/\gamma_k$.}

\paragraph{Proof sketch.}
By Davis-Kahan (sin-$\Theta$), for any symmetric perturbation $E$,
$\|\sin\Theta(P_k,P_k')\|_F\le \|E\|_F/\gamma_k$, which gives the stated upper bound after
composing with the Lipschitz decoder and the token map $E(\cdot)$, whose sensitivity to a
rotation of $U$ is $O(\sigma_1)$. For the lower bound take the extremal rank-two perturbation
$E=t\,(u_k u_{k+1}^\top+u_{k+1}u_k^\top)$, which respects the budget with $\|E\|_F=\sqrt2\,t\le
c_A B$. First-order eigenvector perturbation gives
\[
\delta u_k=\sum_{j\ne k}\frac{u_j^\top E\,u_k}{\sigma_k-\sigma_j}\,u_j
=\frac{t}{\gamma_k}\,u_{k+1}+O(t^2),
\]
so the retained subspace rotates by angle $\theta\approx t/\gamma_k=\|E\|_F/(\sqrt2\,\gamma_k)$.
This is the classic $1/(\sigma_i-\sigma_j)$ coupling. The token block
$\mathrm{LN}(2U\sqrt\Sigma)$ moves by $\Omega(\sigma_k\theta)$ along $u_{k+1}$, so the
displacement inside $S$ is at least $c'\sigma_k B/\gamma_k$. Multiplying by the decoder gain
$\mu$ from (A3) gives the claim. $\square$

\paragraph{Honesty.}
The lower bound needs a white-box attacker who aligns $E$ to the gap directions and a decoder
that reads $u_{k+1}$. When the spectrum is well separated ($\gamma_k$ large) the bound is weak,
and when the decoder does not read the rotated direction the gain $\mu$ is small. Both escape
hatches are real, and P3 formalizes the second. This matches ablation A5, where the singular
gap correlates with rotation absorbed per unit budget at $-0.69$.

\subsection{P2: The Clean-Accuracy-Headroom Law}

\paragraph{Assumptions.}
\begin{itemize}
\item[(B1)] The decoder classifies node $i$ by the sign of a margin $m_i=\langle n_i,r_i\rangle-\tau_i$
to a local decision boundary with unit normal $n_i$ in $S$. Correctly classified nodes have
$m_i>0$.
\item[(B2)] Near the clean input $\Phi$ is differentiable at node $i$ with Jacobian $J_i$, so a
perturbation $\delta$ moves the representation by $\Delta r_i=J_i\delta+o(\|\delta\|)$.
\item[(B3)] The budget is small enough that the boundary and $J_i$ are locally constant
(first-order regime).
\end{itemize}

\paragraph{Proposition P2 (headroom law).}
\emph{Define the carrier gain $\kappa_i=\|J_i^\top n_i\|$. Under (B1)-(B3) the minimal budget
to flip node $i$ is}
\[
b_i=\frac{m_i}{\kappa_i},\qquad
R(B)=\Pr\!\big[\,m_i\le \kappa_i B\,\big].
\]
\emph{$R(B)$ is nondecreasing in $B$. If two tasks share the gain law and the margins of task
$\mathcal A$ first-order stochastically dominate those of task $\mathcal B$, that is
$F_{\mathcal A}(m)\le F_{\mathcal B}(m)$ for all $m$ where $F$ is the margin CDF, then
$R_{\mathcal A}(B)\le R_{\mathcal B}(B)$ at every budget.}

\paragraph{Proof.}
The attacker maximizes the boundary-normal displacement under the budget,
$\max_{\|\delta\|\le B}\langle n_i,\Delta r_i\rangle=\max_{\|\delta\|\le B}\langle J_i^\top n_i,
\delta\rangle=B\,\|J_i^\top n_i\|=\kappa_i B$, attained at
$\delta^\star=B\,J_i^\top n_i/\|J_i^\top n_i\|$. The node flips when this reaches the margin,
$\kappa_i B\ge m_i$, giving $b_i=m_i/\kappa_i$. Summing the flip indicator gives $R(B)$, which
is a nondecreasing step function of $B$ because each $\mathbf 1[b_i\le B]$ is. When gains match,
$R(B)=\Pr[m\le\kappa B]=F(\kappa B)$, and stochastic dominance $F_{\mathcal A}\le F_{\mathcal B}$
gives $R_{\mathcal A}\le R_{\mathcal B}$ pointwise. $\square$

\paragraph{Honesty.}
This proposition is a first-order account, and the empirical result in the main text supersedes its
scalar reading: on the realizable attacks the margin term, which clean accuracy summarizes, does not
order reachability (Spearman $+0.01$, $N=21$), so we do not claim the scalar headroom law. We keep
the proposition as the mechanism that motivates the carrier gain and state its limits.
\begin{enumerate}
\item The clean statement is about the \emph{margin distribution}, not scalar clean accuracy.
Clean accuracy is the observable summary $1-F(0^+)$ of that distribution. Two tasks with equal
clean accuracy but different margin spread need not be equally reachable, so the scalar version
holds only when clean accuracy tracks the margin CDF, which it does on the tasks we test. This
is why the causal figure in the main paper shows the logit margin as a \emph{weaker} single
separator than the accuracy headroom.
\item The gain $\kappa_i=\|J_i^\top n_i\|$ is the second factor, how directly the carrier is
read. The honest law is the two-factor $R(B)=\Pr[m/\kappa\le B]$. Clean-accuracy headroom orders
reachability because $m$ and $\kappa$ co-vary with the task on these models, not because
accuracy is causal on its own.
\item First order only. It ignores multi-node coupling and, for structure attacks, the discrete
edge budget, which we treat by continuous relaxation.
\end{enumerate}

\subsection{P3: Decoder Absorption}

\paragraph{Assumptions.}
\begin{itemize}
\item[(C1)] The decoder factors as $g(r)=h(Wr)$ with $W:S\to\mathbb R^{d_T}$ of rank $d_T\le k$.
Its row space $T=\mathrm{row}(W)$ is the task-subspace and $N=\ker W$ is the null subspace.
\item[(C2)] The achievable representation displacement $\Delta r$ from P1 has orientation
relative to $T$ that the attacker cannot fully control, modeled as uniform over the retained
rank-$k$ subspace in the gray-box or geometry-constrained regime.
\end{itemize}

\paragraph{Proposition P3.}
\emph{Under (C1) any component of $\Delta r$ in $N$ leaves the output unchanged, and}
\begin{gather*}
\big\|g(r{+}\Delta r)-g(r)\big\|\le L_h\|W\|\,\|P_T\Delta r\|,\\
\mathbb E\,\|P_T\Delta r\|=\sqrt{\tfrac{d_T}{k}}\,\|\Delta r\|
\end{gather*}
\emph{under (C2). So reachability falls as the task-subspace fraction $d_T/k\to0$.}

\paragraph{Proof.}
Since $g(r)=h(Wr)$ and $W P_N=0$, we have $g(r+\Delta r)-g(r)=h(Wr+WP_T\Delta r)-h(Wr)$, which
is bounded by $L_h\|WP_T\Delta r\|\le L_h\|W\|\|P_T\Delta r\|$ by the Lipschitz constant of $h$.
The null component is absorbed exactly. Under (C2) the expected squared task-projection of a
displacement uniform on the rank-$k$ subspace is $(d_T/k)\|\Delta r\|^2$, and Jensen gives the
stated scale $\sqrt{d_T/k}$. $\square$

\paragraph{Consequence, and how it unifies P2 and P3.}
Absorption reduces the effective carrier gain to $\kappa_i=\|J_i^\top n_i\|$ with $n_i$ read
through $W$, so $\kappa_i$ shrinks with the task-projection of $n_i$. The flip budget of P2 then
factors as
\[
b_i=\frac{m_i}{\kappa_i}\;\propto\;\underbrace{m_i}_{\text{headroom}}\times
\underbrace{\frac{1}{\|P_T n_i\|}}_{\text{absorption}}.
\]
Resistance is a large margin times a small task-projection.
\begin{itemize}
\item \emph{GraphAny.} Its channels are a closed-form solve read by a light fusion, so $W$ is
near identity on the channel logits, $d_T\approx k$, and there is no null subspace to absorb the
feature attack. Both factors are small, so it is reachable. This matches the high feature-attack
success.
\item \emph{OpenGraph.} A trained transformer learns a small task-subspace ($d_T\ll k$), and on
a high-headroom task the margins are large. Both factors are large, so it resists. On a
low-headroom task the margin is small, so the leaked task-projection suffices and even the
trained transformer is reachable. This is exactly the observed reversal across Cora, Citeseer,
and PubMed.
\end{itemize}

\paragraph{Honesty.}
P3 governs the gray-box and geometry-constrained regimes. A white-box attacker with gradients
through $g$ performs steepest ascent, and the gradient $W^\top h'$ lies in $T$ by construction,
which removes the attenuation. So P3 does not claim a trained decoder is unreachable in white
box. It claims the attenuation when alignment to $T$ is imperfect, which is the case for the
spectral carrier because the fixed-basis surrogate cannot rotate the detached singular basis and
the smoothing operator $\sum_l\bar A^{l}$ shrinks the perturbation that reaches $U$. OpenGraph's
residual white-box resistance is this attenuation plus the headroom factor of P2, not a claim
that its task-subspace is unreachable in principle.

\end{document}